\documentclass[manuscript]{acmart}

\setcitestyle{aysep={,}}

\usepackage[utf8]{inputenc}
\usepackage[T1]{fontenc}
\usepackage{underscore}
\DeclareUnicodeCharacter{03C3}{\ensuremath{\sigma}}

\usepackage{algorithm}
\usepackage{algpseudocode}

\usepackage{graphicx}
\usepackage{subcaption}   
\usepackage{grffile}

\usepackage{booktabs}
\usepackage{tabularx}
\usepackage{array}
\usepackage{multirow}

\usepackage{xcolor}
\usepackage{placeins}
\usepackage{pgffor}
\usepackage{csvsimple}
\usepackage{url}

\begin{document}

\title{Confidence-based Ranking with Adaptive Sampling for Noisy Black-Box Optimisation}

\author{Enrico Halim}
\affiliation{%
  \institution{The University of New South Wales}
  \city{Canberra}
  \country{Australia}
}
\email{enrico.halim@unsw.edu.au}

\author{Hemant Kumar Singh}
\affiliation{%
  \institution{The University of New South Wales}
  \city{Canberra}
  \country{Australia}
}
\email{hemant.singh@unsw.edu.au}

\author{Tapabrata Ray}
\affiliation{%
  \institution{The University of New South Wales}
  \city{Canberra}
  \country{Australia}
}
\email{t.ray@unsw.edu.au}

\renewcommand{\shortauthors}{Halim et al.}

\begin{abstract}
Real-world optimization problems often involve black-box functions and uncertainties in their evaluation, widely referred to as noisy optimization problems~(NOPs). Evolutionary algorithms~(EA), including Evolutionary Strategies~(ES) and genetic algorithms~(GA) have been commonly adopted to solve these problems in the contemporary literature. An ongoing challenge is the computational expense involved, given the number of evaluations required for good fitness estimation and ranking. Two fundamental methods commonly used for fitness estimation for NOPs are \emph{implicit averaging} and \emph{explicit averaging}. Explicit averaging uses resampling of solutions to improve the estimates, while implicit averaging typically uses a large population size, with low resampling. Implicit averaging has been shown to have theoretical advantages for certain cases~(e.g. spherical), which has motivated some state-of-the-art approaches to use them. However, in a recent study, experiments demonstrated that its performance is highly dependent on certain assumptions about the function, such as steepness and constant noise level, which may not apply for majority of the real world problems. Moreover, most existing algorithms~(both implicit and explicit) have been developed and tested under the assumption of \emph{homoscedastic} noise, where the amplitude of variation is uniform across the entire search space, as opposed to more generic case of \emph{heteroscedastic} noise. To address these issues, we introduce a set of heteroscedastic test problems and propose a novel confidence ranking method that employs a computationally efficient explicit averaging strategy with sampling budget adaptation, which allocates the sampling budget toward more promising solutions. The proposed ranking method is implemented within the widely used Covariance Matrix Adaptation ES~(CMA-ES) and canonical GA frameworks to demonstrate its effectiveness and versatility. The resulting algorithm is evaluated on a range of problems with both homoscedastic and heteroscedastic noise, and it demonstrates superior performance compared to state-of-the-art approaches. 
\end{abstract}


\maketitle

\section{Introduction and Related work} \label{sec:intro}

Optimization in practical scenarios is often influenced by uncertainties \cite{jin2005evolutionary, comprehensive_survey}. These uncertainties vary in their sources and their effects on the underlying functions, leading to a few distinct problem classes of research interest, with different targeted solutions and identification approaches.

A common case is when achieving a design variable value precisely is difficult, e.g., due to manufacturing tolerances. Such variations propagate to objectives and constraints. The goal in this scenario is to find solutions whose performance remains resilient to these unavoidable variations. This is known as \emph{robust optimization} where the focus is on variations in the objective values \cite{six_sigma_design,beyer2007robust}. In the similar paradigm, \emph{reliability optimization} is typically focused on satisfaction of constraints under uncertainties \cite{deb2009reliability}.

Another case involves uncertainties in estimating the objective function itself, which may result from measurement noise, modeling errors, or stochastic processes. This is referred to as noisy optimization problem (NOP). The aim in this scenario is to identify the optimum of the true (typically mean or expected) function without the noise~\cite{OPLCMA,revisiting,comprehensive_survey}. This is particularly challenging in black-box settings, where the objective and noise models are not separately known and must be inferred through sampling. A related scenario occurs when constraints themselves involve uncertainty. In such cases, solutions must satisfy constraints with a certain probability. These are known as chance-constrained optimization problems \cite{neumann2025runtime}. Evidently, a single problem may involve multiple types of uncertainties simultaneously. For a comprehensive review of optimization under uncertainty, readers are referred to surveys such as \cite{beyer2007robust,comprehensive_survey,jin2005evolutionary}. 

In this study, we focus on unconstrained single-objective NOPs. As noted above, in black-box settings, the true (noiseless) objective function \(f(x)\) cannot usually be evaluated exactly. Instead, each evaluation produces a noisy or stochastic observation, denoted as \(\hat f(x)\), which is modeled as:

     \begin{equation}
      \hat f(x) \;=\; f(x) + \varepsilon
    \end{equation}
    
The noise $\varepsilon$ can generally be classified into two categories:

\begin{itemize}
    \item \textbf{Homoscedastic:} This is when the variance of the noise is independent of the location in the search space \cite{homos_def}. It is defined as:
\begin{equation}
  \varepsilon \sim \mathcal{N}(0,\sigma^2)
  \quad\Longrightarrow\quad
  \hat f(x)\sim\mathcal{N}\bigl(f(x),\,\sigma^2\bigr)
\end{equation}

Here, \(\varepsilon\) follows a normal distribution with zero mean and constant standard deviation \(\sigma\). This formulation usually captures measurement noise.

\item \textbf{Heteroscedastic:} This represents the more general case, where the noise level varies across the search space~\cite{hetero_def}. For example, in simulation-based optimization, certain regions of the decision space may correspond to unstable or highly sensitive system behavior, leading to larger variability in the observed objective values, while other regions produce more stable and consistent evaluations. It is defined as: 
\begin{equation}
  \varepsilon \sim \mathcal{N}(0,g(x))
  \quad\Longrightarrow\quad
  \hat f(x)\sim\mathcal{N}\bigl(f(x),\,g(x))
\end{equation}
Here, the variance of the noise depends on the location $x$. Two common models include:
\begin{itemize}
    \item Fitness-proportional noise: $g(x)=c\,|f(x)|$, where larger objective values correspond to larger noise ~\cite{fitness_proportionate}. 
    \item Inverse-proportional noise: $g(x)=c/|f(x)|$, where larger objective values correspond to smaller noise ~\cite{worst_best_noise}. 
\end{itemize}
\end{itemize}

Although in this paper we assume that $\varepsilon$ follows a normal distribution~(consistent with large body of literature in the domain), noise in practical applications is not limited to the Gaussian case, and other distributions may arise depending on the system.

\subsection {Evolutionary approaches for noisy optimization}
A comprehensive review of EAs for solving noisy optimization problems can be found in \cite{comprehensive_survey}. 
An essential component of any EA designed to address noise is its strategy for improving the fitness estimates of explored solutions, 
as these estimates directly influence ranking and selection methods. In this context, two notable approaches are commonly employed: \emph{explicit averaging} and \emph{implicit averaging}.

\begin{itemize}
    \item {\textbf{Explicit averaging:}}
This strategy repeatedly evaluates the stochastic objective function of a candidate solution multiple times, and uses the sample mean as its fitness estimate~\cite{explicit_def}. Statistically, $n$ repeated evaluations of an individual’s objective function reduce the standard error of the sample mean by a factor of $\sqrt{n}$~\cite{sqrt_n_error}. Explicit averaging strategy can be further classified into static and dynamic sampling~\cite{comprehensive_survey}. In static sampling, a constant sample size is allocated across all evaluations. In contrast, dynamic sampling adjusts the sample size adaptively. For example, more samples may be allocated to solutions with larger variance (standard deviation) compared to those with smaller variance (standard deviation)~\cite{OCBA}. Alternatively, fewer samples may be used in the early stages of the run, with the number progressively increased to enable more accurate selection towards the end of the run, thereby improving fine-tuning in near-optimal regions~\cite{incremental_sampling}.

\item {\textbf{Implicit averaging:}} In contrast to explicit averaging, this strategy does not attempt to evaluate a candidate solution multiple times. Instead, it combats noise through the use of relatively larger population sizes. For example, in ~\cite{revisiting}, a fixed evaluation budget of $1024$ function evaluations per iteration was considered. Under this setting, pure implicit averaging evaluates $1024$ distinct candidate solutions once per iteration, whereas pure explicit averaging uses the minimum population size of $4$ and evaluates each candidate solution $256$ times per iteration. The underlying idea is that population-based metaheuristics naturally revisit the same or similar candidate solutions across generations, thereby implicitly averaging out their noisy evaluations \cite{comprehensive_survey,OPLCMA,revisiting}.

\end{itemize}

Evidently, both explicit averaging and implicit averaging require a significant number of evaluations, 
particularly when employed within EA frameworks. Consequently, there is ongoing research interest in improving these sampling techniques~\cite{comprehensive_survey}. 
At the same time, several studies have sought to broadly analyze and compare explicit and implicit averaging strategies to determine which is more computationally efficient~\cite{revisiting,comprehensive_survey,beyer2007evolutionary}. In this regard, a few theoretical investigations have highlighted the advantages of implicit averaging over explicit averaging~\cite{arnold2006general,arnold2001local}, especially for spherical and quadratic functions. This has motivated the adoption of implicit averaging in a number of recent evolutionary approaches for solving noisy optimization problems~\cite{hansen2009benchmarking,OPLCMA,hellwig2016evolution}. 

However, a recent study~\cite{revisiting} systematically examined implicit and explicit averaging under various experimental settings and the findings indicate that advantages of implicit methods depend on problems exhibiting specific properties, which may not hold generally across a diverse range of problems. These properties include steepness of the objective function, symmetry about the global optimum, and homoscedastic noise. Such reliance on problem-specific features suggests that implicit averaging may be unsuitable as a general-purpose solver for noisy optimization problems.

Recent developments in noisy black-box optimization have frequently adopted CMA-ES as the baseline framework. Variants of CMA-ES designed for noisy optimization problems (NOPs) differ primarily in how they handle noise. Four representative methods are OPL-CMA-ES~\cite{OPLCMA}, PSA-CMA-ES~\cite{PSA-CMA-ES}, LRA-CMA-ES~\cite{LRA-CMA-ES}, and RA-CMA-ES~\cite{RA-CMA-ES}. 

PSA-CMA-ES~\cite{PSA-CMA-ES} and OPL-CMA-ES~\cite{OPLCMA} both rely on implicit averaging by adapting the population size of CMA-ES. In PSA-CMA-ES, the population size is adjusted based on the reliability of the distribution update, estimated from the length and consistency of an evolution path in the distribution-parameter space. A long, stable evolution path indicates reliable updates and allows the population size to remain small, whereas short or inconsistent paths trigger population enlargement to obtain a more accurate evolution. OPL-CMA-ES also uses population-size adaptation strategy, but estimates noise strength more directly by reevaluating the population and measuring rank changes. Large rank fluctuations indicate that noise is disrupting selection, prompting an increase in population size. Both methods improve robustness, but their primary response to noise remains population enlargement.

LRA-CMA-ES~\cite{LRA-CMA-ES} takes a different approach by keeping the population size fixed at the default CMA-ES value and instead adapting the learning rates of the distribution updates. The learning rate determines how strongly the estimated update direction influences the mean vector and covariance matrix. Larger learning rates accelerate progress but can amplify unreliable updates, while smaller learning rates yield more conservative and stable search behavior. LRA-CMA-ES estimates the signal-to-noise ratio of the update direction and adjusts the learning rates to maintain this ratio near a target value.

RA-CMA-ES focuses on explicit averaging strategy by adapting the number of re-evaluations assigned to each candidate solution~\cite{RA-CMA-ES}. The method is motivated by the observation that population-size and learning-rate adaptation may be unreliable under certain noise structures, particularly multiplicative noise. RA-CMA-ES estimates the reliability of the update direction by splitting the evaluations into two groups, computing two update directions, and measuring their correlation. If the two directions are poorly correlated, the number of re-evaluations is increased; otherwise, it can be reduced. This improves robustness under multiplicative noise while maintaining competitive performance under homoscedastic noise. Nevertheless, repeated evaluations can still be expensive, particularly when many samples are required to obtain a reliable estimate.

Motivated by the observation that population-size and learning-rate adaptation may be unreliable under multiplicative noise, RA-CMA-ES~\cite{RA-CMA-ES} focuses on explicit averaging and adapting the number of re-evaluations in each generation. RA-CMA-ES splits evaluations into two groups, computes two update directions, and measures their correlation. Poorly correlated directions indicate unreliable updates and trigger an increase in the number of re-evaluations. On the other hand, strong correlation allows the number of samples to be reduced. This approach improves robustness under multiplicative noise while maintaining competitive performance under homoscedastic noise. However, repeated evaluations can still be expensive when many samples are required to obtain a reliable estimate.

Overall, recent CMA-ES variants enhance noise robustness by adapting the population size, learning rate, or number of re-evaluations. However, these mechanisms are primarily designed to safeguard the reliability of CMA-ES distribution updates, making them heavily tailored to CMA-ES and less transferable to other metaheuristics. Moreover, they can still be inefficient in computationally expensive settings because evaluation budgets are consumed through larger populations, slower population movement, or repeated sampling across all candidates. These broad budget allocation strategies does not prioritize evaluations for candidate solutions that are most critical for selection or in determining the best solution. 

\subsection{Motivation and contributions}

Following from the findings of \cite{revisiting}, the key aim of this work is to design a computationally efficient \emph{explicit averaging} approach that is competitive with current available methods. The motivation is two-fold. First, explicit averaging has greater potential for generalization to problems where the noise profile does not conform to the specific assumptions discussed in the previous section. Second, explicit averaging provides a stronger basis for integration into other optimization frameworks, including those that are not population-based. For example, in surrogate-assisted or Bayesian optimization~\cite{explicit_for_surrogates}, a large number of solutions may not be available in a given neighborhood, unlike in population-based methods. In such cases, implicit approaches may be inherently unsuitable for guiding the search. Although the development of new surrogate-assisted approaches is beyond the scope of this study, the techniques proposed here could be extended to such frameworks in future work.

To address the research gaps identified in the discussions above, we present two key contributions in this work: 

\begin{itemize}
    \item First, we propose a novel confidence ranking (CR) method that employs a computationally efficient explicit averaging strategy with sampling budget adaptation. This strategy allocates a larger sampling budget to more promising solutions while reducing resources devoted to poorer ones. Using this method, an algorithm can dynamically terminate resampling once the desired confidence level in the ranking is achieved. The proposed ranking method is subsequently implemented within the CMA-ES framework, referred to as CR-CMA-ES, and within the GA framework, referred to as CR-EA, to demonstrate its performance. To the authors' knowledge, similar confidence-based selection has previously been explored only in \cite{CDR}, where it was specifically designed for multi-objective optimization (e.g., using crowding distance as a selection criterion). No equivalent approaches have been proposed for single-objective noisy optimization problems, as well as no in-depth study is available with regards to search behaviour and their variation across \emph{homoscedastic} and \emph{heteroscedastic} problems.

    \item The second major contribution of this work is the experimental evaluation of the compared techniques across heteroscedastic benchmark problems, in addition to the conventionally used homoscedastic problems. To this end, five different types of noise profiles are constructed and combined with various base functions. Additionally, different level/intensities of noise amplitude are considered. Additionally, we also illustrate the performance on simulation-bsaed applications. In contrast, most studies in noisy optimization focus exclusively on homoscedastic test cases~\cite{comprehensive_survey}. This study therefore aims to provide a more comprehensive assessment of algorithmic performance. Furthermore, several instances of simulation-based noisy problems from the domain of chemistry are also examined. The results indicate that CR-CMA-ES and CR-EA outperform their respective CMA-ES and EA variants.
\end{itemize}

In Section~\ref{sec:approach}, we present the details of the proposed method, followed by a discussion of the experimental setup in Section~\ref{sec:expsetup}. Results on benchmark problems are discussed in Section~\ref{sec:results}, followed by simulation-based applications in Section~\ref{sec:olympus}. Concluding remarks and potential future research directions are provided in Section~\ref{sec:conc}.

\section{Proposed Method}
\label{sec:approach}
This section provides a detailed description of the proposed method and its implementation. Throughout this study, computational cost is measured in terms of the number of function evaluations, which is assumed to be the dominant computational expense. This is often the case in many practical~(especially engineering) scenarios due to the use of simulations or physical experiments. Section~\ref{ranking method} first introduces the three key components of the proposed approach: a novel ranking scheme, an adaptive resampling strategy, and sampling budget adaptation. The proposed approach is generic enough that it can be integrated in different commonly used optimization frameworks to improve their performance. To demonstrate this modularity of the approach, we integrate it into two different algorithmic frameworks, namely CMA-ES and GA, as described in Section~\ref{ranking implementation}. CMA-ES is selected because many recent algorithmic developments in the literature build upon this framework, while GA is included to represent evolutionary approaches that employ elitism. 

\subsection{Confidence Ranking with Budget Adaptation}\label{ranking method}
This is a general-purpose ranking method applicable to any algorithm that relies on ranking candidate solutions as part of its search strategy. The method operates by repeatedly evaluating the objective value of candidate solutions produced by the algorithm and using the resulting outcomes to construct a ranking that the algorithm can subsequently use.

\subsubsection{\textbf{Initial Sampling}}
Each candidate solution $x_i$ in the pool of size $N$ is first evaluated with a minimum sample size $s_0$, and the corresponding estimates $\hat{f}_i$, along with $x_i$, are stored in an evaluation archive. This archive is continually updated whenever any solution is evaluated throughout the algorithm run. 

A user-defined parameter $S_t$ prescribes the global sampling budget for generation $t$ as: 
\begin{equation}
    B_t = N \times S_t.
\end{equation}
If all solutions were to be evaluated with the same number of samples, each would receive $S_t$. However, instead of uniform sampling, we assign a maximum sample size to each solution such that the average equals $S_t$. 

The maximum sample size $s_x$ is determined in proportion to the standard deviation of the solutions, as shown in Equation~(\ref{eq:sol-max-sample}). 

\begin{equation}
s_x = B_t \times
      \frac{\sigma_{x_i}}{\sum_{{x_i} \in X}\sigma_{x_i}}.
\label{eq:sol-max-sample}
\end{equation}
Here, $X$ is the set of solutions in the current pool, $\sigma_{x_i}$ is the sample standard deviation of solution $x_i$, and $\sum_{x_i \in X}\sigma_{x_i}$ is the total sample standard deviation across the current pool. From the equation, it can be inferred that this allocation distributes more samples to high-noise regions and fewer samples to low-noise regions. This is consistent with recommendations in Optimal Computing Budget Allocation (OCBA) studies~\cite{OCBA}, which have established that assigning samples proportionally to the standard deviations helps maximize the probability of correctly selecting the best solution in stochastic simulations. The sampling budget prescribed through this process is utilized later in the ranking procedure as will be discussed shortly. 

\subsubsection{\textbf{Ranking Procedure}}
Each solution in the pool is then compared against all others (all possible pairwise comparisons are performed) using Welch’s $t$--test at a $C$ confidence level ($C$ is a user-
defined parameter). Welch’s $t$--test is chosen due to its ability to deal with normal distributions with unequal variances and its robustness against Type-I error~\cite{derrick2016welch}. For each solution, $f_{\mathrm{win}}$, $f_{\mathrm{loss}}$, and $f_{\mathrm{tie}}$ are initialized to zero prior to the comparisons. During each pairwise comparison, $f_{\mathrm{win}}$ is incremented by one if the solution is significantly better than its counterpart, $f_{\mathrm{loss}}$ is incremented by one if it is significantly worse, and $f_{\mathrm{tie}}$ is incremented by one otherwise. After all comparisons, candidates are ranked lexicographically in descending order of  $(f_{\mathrm{win}}, f_{\mathrm{tie}}, -f_{\mathrm{loss}})$.  The top-ranked solution is designated as the proposed best solution of the generation.

Note that the choice of statistical operator is flexible and can be appropriately selected by a user to suit anticipated noise profiles~(if known). In this work, Welch's $t$-test is used for demonstration purposes because it is widely known and relatively easy to interpret. Also to note, when ranking the population members, the Welch’s $t$-test is applied pairwise, which naturally introduces the possibility of family-wise error. We have considered the potential impact of this issue. However, the algorithm does not require a perfectly accurate global ranking of all individuals; it only needs to reliably maintain a sufficiently strong set of survivors to guide the search effectively. We further support this design choice with additional plots presented in section~\ref{sec:true_rank}. 

\subsubsection{\textbf{Resampling Procedure}}
Following the initial ranking described in the previous section, the algorithm identifies the set of solutions that are statistically indistinguishable from the top $\gamma$-ranked solution according to Welch’s $t$--test. These are referred to as critical top-ranked solutions. Specifically, it aggregates the unique $f_{\mathrm{tie}}$ partners associated with each of the top $\gamma$-ranked solutions, where $\gamma$ is a user-defined parameter that can be set based on the type of framework used~(reasoning for $\gamma$ values used for CMA-ES and GA are discussed later in numerical experiments).

Among the solutions identified above, those that have not yet reached their maximum sample size $s_x$ are selected. This set is then combined with the current top $\gamma$-ranked solution that has also not reached its maximum sample size. The resulting combined set, referred to as \emph{evaluation candidates}, receives additional evaluations. The number of additional samples allocated is determined as per Equation~(\ref{eq:add_samples}).

\begin{equation}
s_{+} = \max\!\left(\Bigl\lfloor \tfrac{s_x}{S_t}\Bigr\rfloor, \, 1\right).
\label {eq:add_samples}    
\end{equation}

Intuitively, incrementing the sampling by one evaluation at a time is desirable. However, this may slow the runtime when the allowed sampling budget $s_x$ is large. Therefore, to keep the number of resampling loops relatively low (approximately $S_t$), we employ Equation~(\ref{eq:add_samples}). 

Since the standard deviation $\sigma_x$ of a given solution changes after each resampling, the maximum sample size $s_x$ is updated according to Equation~(\ref{eq:sol-max-sample}) after every resampling step. 

The ranking and resampling procedures discussed above are repeated until one of the following conditions is satisfied:  
(a) no solution in the pool remains tied with the top $\gamma$-ranked solution, or  
(b) there is no more \emph{evaluation candidates}. 

\subsubsection{\textbf{Sampling Budget Adaptation}}
\label{sec: sampling budget update}

\begin{figure}[!ht]
  \includegraphics[width=0.48\textwidth]{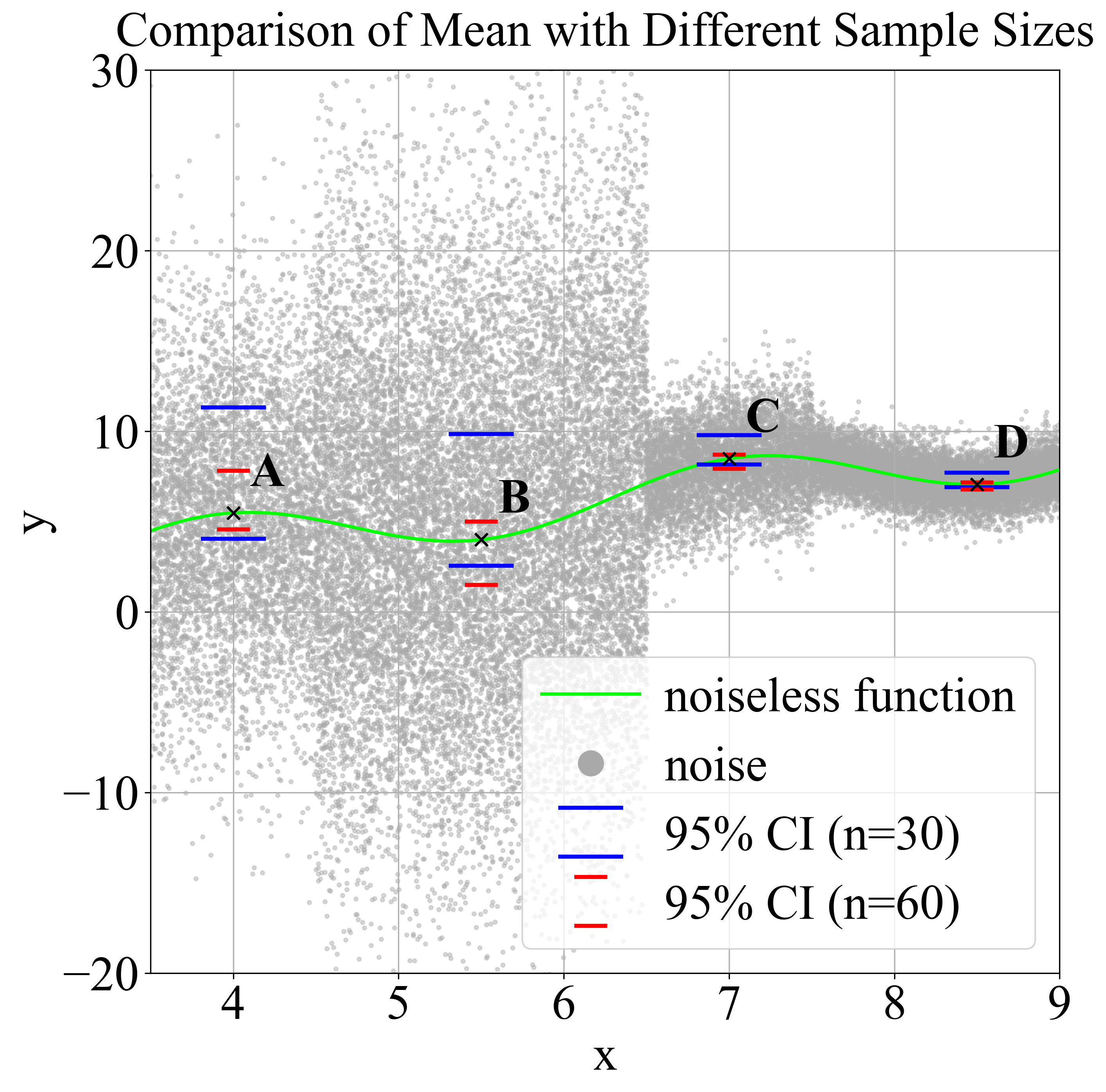}

  \caption{Representative scenario where a higher budget needed is required for correct ranking. The noiseless function is $y = 1.5\sin(2x) + x$. It is subject to Gaussian noise with $mean=y$  and standard deviations of $\sigma = 6$ for $3.5 \leq x \leq 4.5$, $\sigma = 9$ for $4.5 < x \leq 6.5$, $\sigma = 2$ for $6.5 < x \leq 7.5$, and $\sigma = 1$ for $7.5 < x \leq 10$.}
  
  \label{fig:affect of noise}
\end{figure}

Varying noise in the search space can cause the population to incorrectly identify the top-ranked solution(s). Figure~\ref{fig:affect of noise} illustrates four points, labelled A, B, C, and D on a noisy function. The black cross denotes the corresponding position on the noiseless function. 
The blue and red bars represent the $95\%$ confidence intervals obtained using samples of size $30$ and $60$, respectively.

As illustrated in Figure \ref{fig:affect of noise}, pairwise comparisons using Welch’s t–test with 30 samples may result in ties for most solution pairs, except for the combination of solutions C and D, where D marginally outperforms C. Consequently, D is mistakenly selected as the best solution for that generation. This reduces selection pressure in the population and leads to stagnation in convergence. However, when the sample size is increased to 60, the likelihood of correctly identifying solution B as the true top-ranked solution also increases.

This effect is even more prominent as in later generations of the search algorithm because as the population converges, solutions are closer to each other. In this scenario, the number of solutions that are tied with the top ranked solution will increase substantially, even when they have been evaluated with the maximum allowed samples~($s_x$).

To address this issue, the proposed approach incorporates a dynamic adjustment mechanism for the global sampling budget. The mechanism enables the algorithm to adaptively respond to noise levels observed in the ranking process. When a large proportion of the population ends up tied with the top $\gamma$-ranked solution at the end of a generation, it indicates that noise has a strong influence on rankings. In such cases, the global sampling ~($B_t$) is increased to improve confidence in the ranking outcomes. Formally, at the end of each generation, if the number of unique solutions tied with the top $\gamma$-ranked solution exceeds $\beta N$, then the global maximum sampling budget $B_t$ is scaled by a multiplier $\alpha$  ($\beta$ and $\alpha$ are user-defined parameters). 


\subsection{Implementation in CMA-ES and GA frameworks} \label{ranking implementation}

This section describes the implementation of the proposed ranking framework within the CMA-ES and GA optimisation frameworks. Although the same CR framework is employed in both algorithms, the implementation and recommended parameter settings differ slightly because of the distinct characteristics of the underlying search methods. CMA-ES is a non-elitist evolutionary strategy that typically operates with a relatively small population, where the entire population is replaced at each generation. Consequently, ranking errors can have a substantial impact on the search process, requiring more conservative parameter settings. In contrast, the GA implementation is elitist and generally employs a larger population, allowing high-quality solutions to survive across generations. This makes the search less sensitive to occasional ranking errors and permits more aggressive parameter settings. The following first provides general guidelines for selecting the ranking parameters, after which the specific implementations of CR-CMA-ES and CR-EA are presented. 

The parameter settings for $C$, $\gamma$, $\beta$, and $\alpha$ may need to be adjusted depending on the population size and how the chosen algorithm utilizes ranking information. A larger value of $C$ reduces the probability of a Type~I error, but at the cost of increased sampling effort. The parameter $\gamma$ determines how many top-ranked solutions are identified with higher certainty in each generation. When the population size is small, the algorithm has limited information available. Consequently, $\gamma$ typically needs to be set to a higher value, as even a small number of ranking errors can severely disrupt the search process. In contrast, for larger population sizes, $\gamma$ can be set as low as 1, since implicit averaging effects begin to emerge and assist the search process. The parameters $\beta$ and $\alpha$ control how aggressively the sampling budget is adapted relative to the noise level. Care should be taken not to set $\beta$ too low or $\alpha$ too high, as this may cause the sampling budget per generation to increase substantially, prior to the population reaching a promising region. Based on empirical studies, the authors recommend using \(\beta \geq 0.2N\) and \(\alpha \leq 2\) to maintain a reasonable balance between exploration and sampling effort. Recommended parameter settings for CR-EA and CR-CMA-ES are presented in the following subsections.

\subsubsection{\textbf{CR-CMA-ES}}
\label{sec:method_cma}
\begin{figure}[!ht]
  \includegraphics[width=0.90\textwidth]{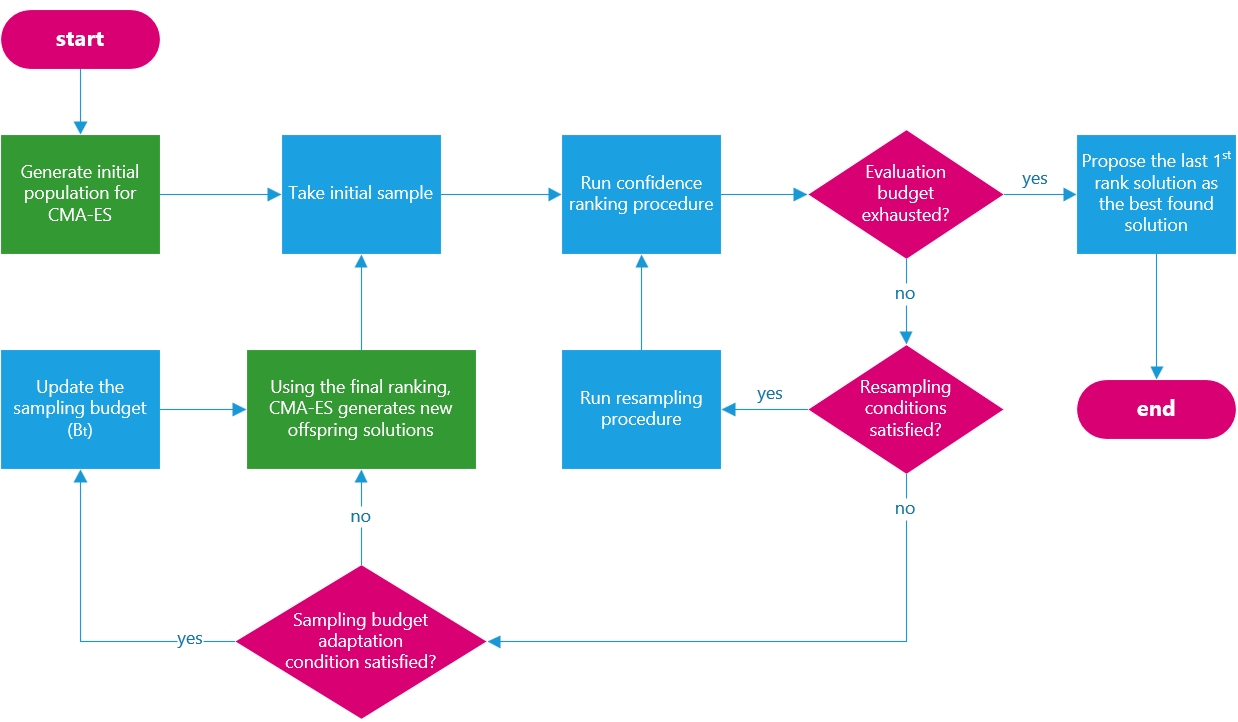}
\caption{Flowchart of CR-CMA-ES. The green boxes correspond to standard CMA-ES components, whereas the other elements correspond to the novel mechanisms introduced in this work.}
  \label{fig:flowchart_cma}
\end{figure}

The implementation in CMA-ES, referred to as CR-CMA-ES, is based on the open-source library~\cite{pycma}. The algorithm is executed using default settings with \texttt{noise\_handler=False}. Two variants are considered: an unbounded version (CR-CMA-ES-UB) and a bounded version (CR-CMA-ES-B). The only modification is the incorporation of the proposed CR method with adaptive sampling budget allocation. The integration of this mechanism is illustrated in the flowchart shown in Figure~\ref{fig:flowchart_cma}. 

The parameter values of $\gamma = 0.5N$ and $\beta = 0.5N$ are chosen for CR-CMA-ES to align with the default CMA-ES setting, which selects the top $0.5N$ ranked solutions as parent solutions. In addition, the $\alpha$ value is set to 1.2 empirically~(the rationale for this setting will be discussed in Section~\ref{sec: abla}). 

\subsubsection{\textbf{CR-EA}}
\begin{figure}[!ht]
  \includegraphics[width=0.90\textwidth]{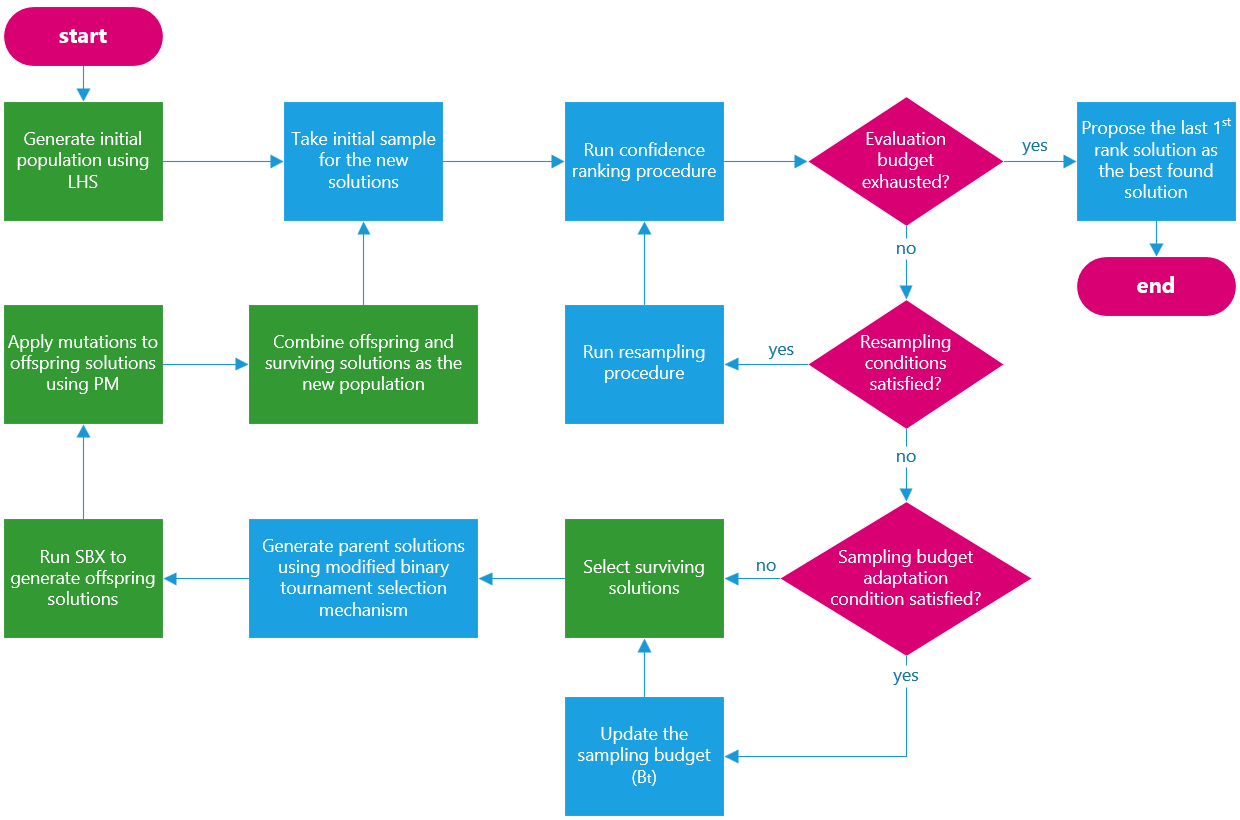}
\caption{Flowchart of CR-EA. The green boxes correspond to standard GA components, whereas the other elements correspond to the novel mechanisms introduced in this work.}
  \label{fig:flowchart_ga}
\end{figure}

The implementation in a canonical GA framework, referred to as CR-EA, is initialized using Latin Hypercube Sampling (LHS). Offspring generation employs widely used evolutionary operators, namely simulated binary crossover (SBX)~\cite{nsga2} and polynomial mutation (PM)~\cite{nsga2}. The $N$ parent solutions are selected using a modified binary tournament selection mechanism adapted to account for uncertainty:
\begin{enumerate}
    \item Two candidates are randomly drawn from the current population.
    \item They are compared using Welch’s $t$--test at the 95\% confidence level.
    \item The candidate that is significantly better is selected as a parent; if no significant difference is detected, one solution is chosen at random with equal probability.
\end{enumerate}
The CR procedure with adaptive sampling budget allocation is then applied to the pool of $2N$ solutions, consisting of both surviving and offspring solutions. An overview of the novel mechanism integration to GA is illustrated in Figure~\ref{fig:flowchart_ga}.

For CR-EA, the $\gamma$ value is set to 1, which means only the first-ranked solutions will be considered as the critical top rank solution, whereas for CR-CMA-ES it is set to $\gamma = 0.5N$. This difference in $\gamma$ value is motivated by the population characteristics of the two algorithms. CR-EA employs a substantially larger population size compared to CR-CMA-ES, making the evolutionary process less sensitive to perfectly ranking the entire population. The larger population in CR-EA also indirectly introduces an implicit averaging effect, which helps stabilise the search under noisy conditions. Furthermore, while the algorithm attempts to confidently identify the first-ranked solution, additional samples are naturally allocated to other solutions with similar performance, which in turn improves the ranking accuracy of nearby solutions as well. The use of binary tournament selection based on Welch’s $t$-test also helps mitigate the effect of imperfect sampling and ranking inaccuracies during evolution. In addition, the value of $\alpha$ and $\beta$ is set to 2 and 0.2 respectively (the rationale for this setting will be discussed in Section~\ref{sec: abla}). 

\section{Numerical Experiments}
\label{sec:expsetup}

\subsection{Benchmark and Noise functions}
\label{sec:benchmark problem}
A total of 100 noisy optimization instances are considered to benchmark the proposed approach against peer methods. These 100 instances are obtained by combining 5 unimodal and 5 multimodal benchmark problems with 5 noise models and 2 noise levels/intensities. All benchmark problems used in this study are minimization problems and defined in Table~\ref{tab:function_settings}, while the noise profiles are outlined in Table~\ref{tab:noise_equations}. The parameter noise level ($c$) is set to $0.3$ and $1.0$, corresponding to the low-noise and high-noise settings, respectively.

\begin{table}[!ht]
  \centering
  \caption{Benchmark functions (all problems are 10 dimension with $f_{\mathrm{opt}} = 0$. $\bar m$ is a normalization factor).}
  \label{tab:function_settings}
  \setlength{\tabcolsep}{6pt} 
  \begin{tabular}{@{}l l r r@{}}
    \toprule
    Problem            & Modality & Bounds        & $\bar m$ \\
    \midrule
    fbenign\_ellipsoid \cite{OPLCMA} & Uni   & $[-50, 100]^{10}$     & 93767 \\
    fdiffpow \cite{OPLCMA}           & Uni   & $[-50, 100]^{10}$     & 327850013219130 \\
    fcigar \cite{OPLCMA}             & Uni   & $[-50, 100]^{10}$     & 5766201855 \\
    ftablet \cite{OPLCMA}            & Uni   & $[-50, 100]^{10}$     & 1040184 \\
    frosenbrock \cite{OPLCMA}        & Uni   & $[-50, 100]^{10}$     & 826496405 \\
    frastrigin \cite{OPLCMA}         & Multi & $[-50, 100]^{10}$     & 7015 \\
    fackley \cite{OPLCMA}            & Multi & $[-50, 100]^{10}$     & 21 \\
    fweierstrass \cite{CEC2014}      & Multi & $[-200, 400]^{10}$ & 14 \\
    fgriewank \cite{Griewank}        & Multi & $[-600, 1200]^{10}$ & 270 \\
    flevy \cite{levy}                & Multi & $[-50, 100]^{10}$   & 1695 \\
    \bottomrule
  \end{tabular}
\end{table}
   
   \begin{table}[!ht]
      \centering
      \caption{Heteroscedastic noise profiles}
      \label{tab:noise_equations}
      \begin{tabular}{lll}
        \toprule
        Noise type               & Equation ($g(x)$)                  & $c$ values                \\
        \midrule
        homoscedastic            & $c$              & $\{0.3,\,1.0\}$  \\
        fitness\_proportional    & $c\,f(x)$              & $\{0.3,\,1.0\}$  \\
        inverse\_proportional    & $c\,|1 - f(x)|$                    & $\{0.3,\,1.0\}$  \\
        middle\_proportional     & $c\,\lvert 0.5 - f(x)\rvert$ & $\{0.3,\,1.0\}$  \\
        sin\_proportional        & $\frac{c\sin\bigl(15\,f(x)-\tfrac{\pi}{2}\bigr)+1}{2}$
        & $\{0.3,\,1.0\}$ \\
        \bottomrule
      \end{tabular}
    \end{table}
   
    To ensure that each noise model has a comparable impact across all benchmark functions, we would like the typical first generation's best solution (in the absence of noise) to $F(x) \approx 1$. This ensures that each noise profile affects all ten benchmark functions on roughly the same normalized scale. To achieve this, we normalize the search space by first taking 100 sets of $N$ points using LHS. Denoting $\mathbf{x}_{i,j}$ as the $j^{th}$ solution of $i^{th}$ set, we compute the average minimum value across the sets as:
    
    \begin{equation}
        \bar m  = \frac{1}{100} \sum_{i=1}^{100} \min_{1 \le j \le N} f(\mathbf{x}_{i,j})
    \end{equation}

    We then use $\bar m$ to transform the search space as:
    \begin{equation}
        F(x) = \frac{f(x)-f_{\mathrm{opt}}}{\bar m -f_{\mathrm{opt}}}
    \end{equation}

    For reproducibility in future studies, the value of $\bar m$ determined through the above process is listed in Table~\ref{tab:function_settings}. Note that this computation is done only once and is external to the algorithm runs. Thus this does not add to the evaluation count for any of the algorithms.

\subsection{Noise Model Illustration}
\begin{figure*}[!htbp]
  \centering
  \subfloat[Homoscedastic noise]
    {\includegraphics[width=0.32\textwidth]{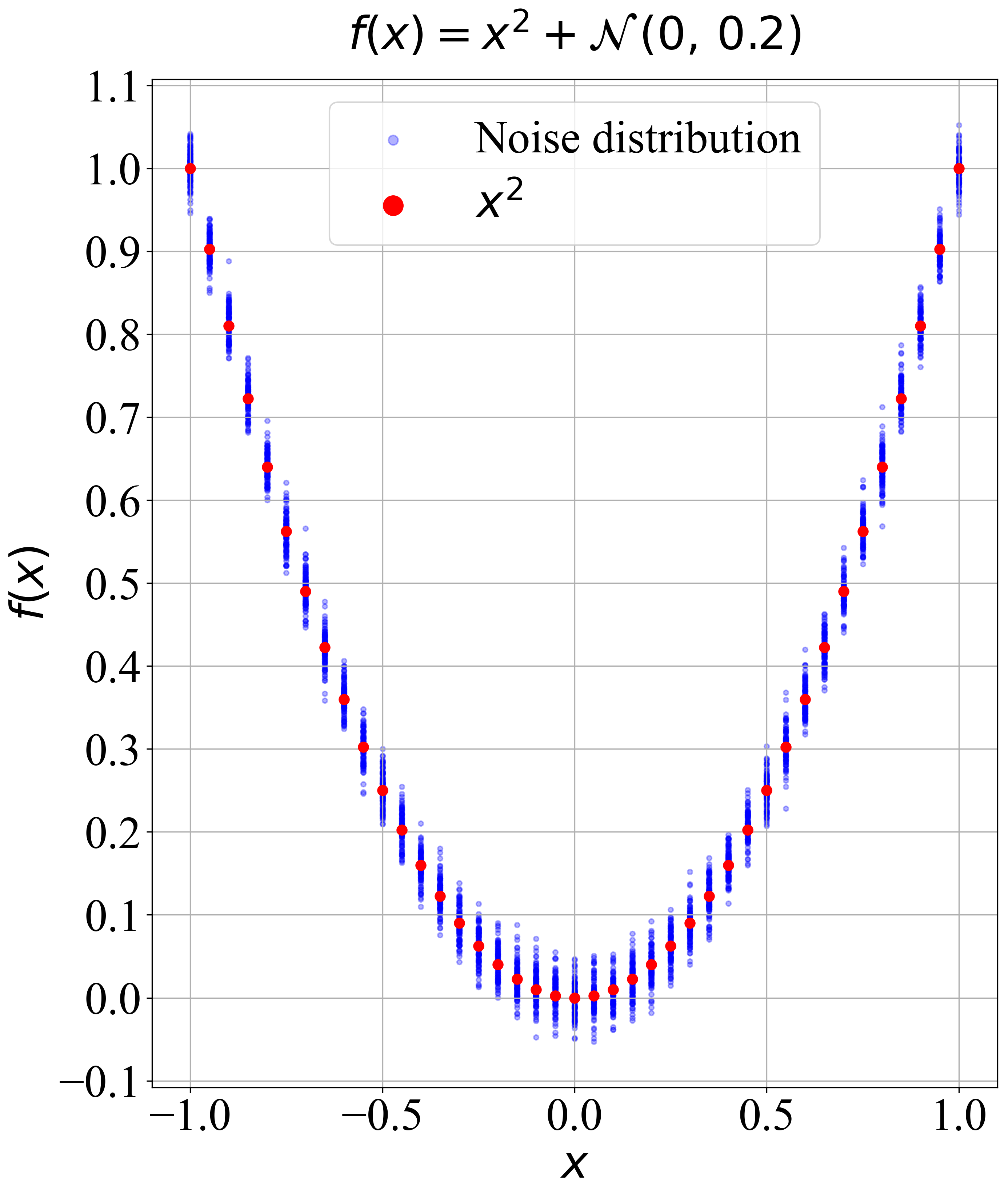}%
     \label{fig:constant_noise}}
  \hfil
  \subfloat[Fitness proportional noise]
    {\includegraphics[width=0.32\textwidth]{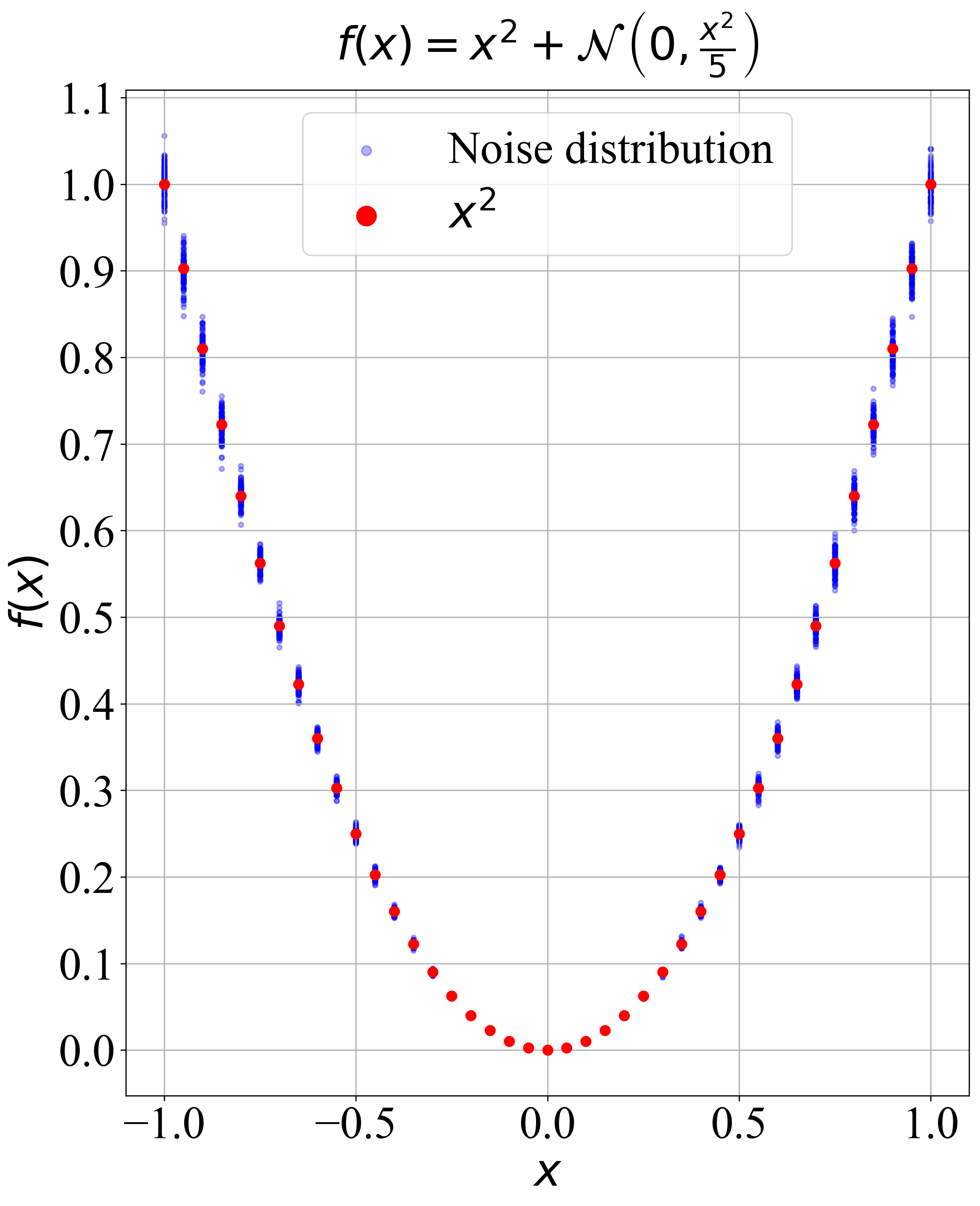}%
     \label{fig:fitness_noise}}
  \hfil
  \subfloat[Inverse proportional noise]
    {\includegraphics[width=0.32\textwidth]{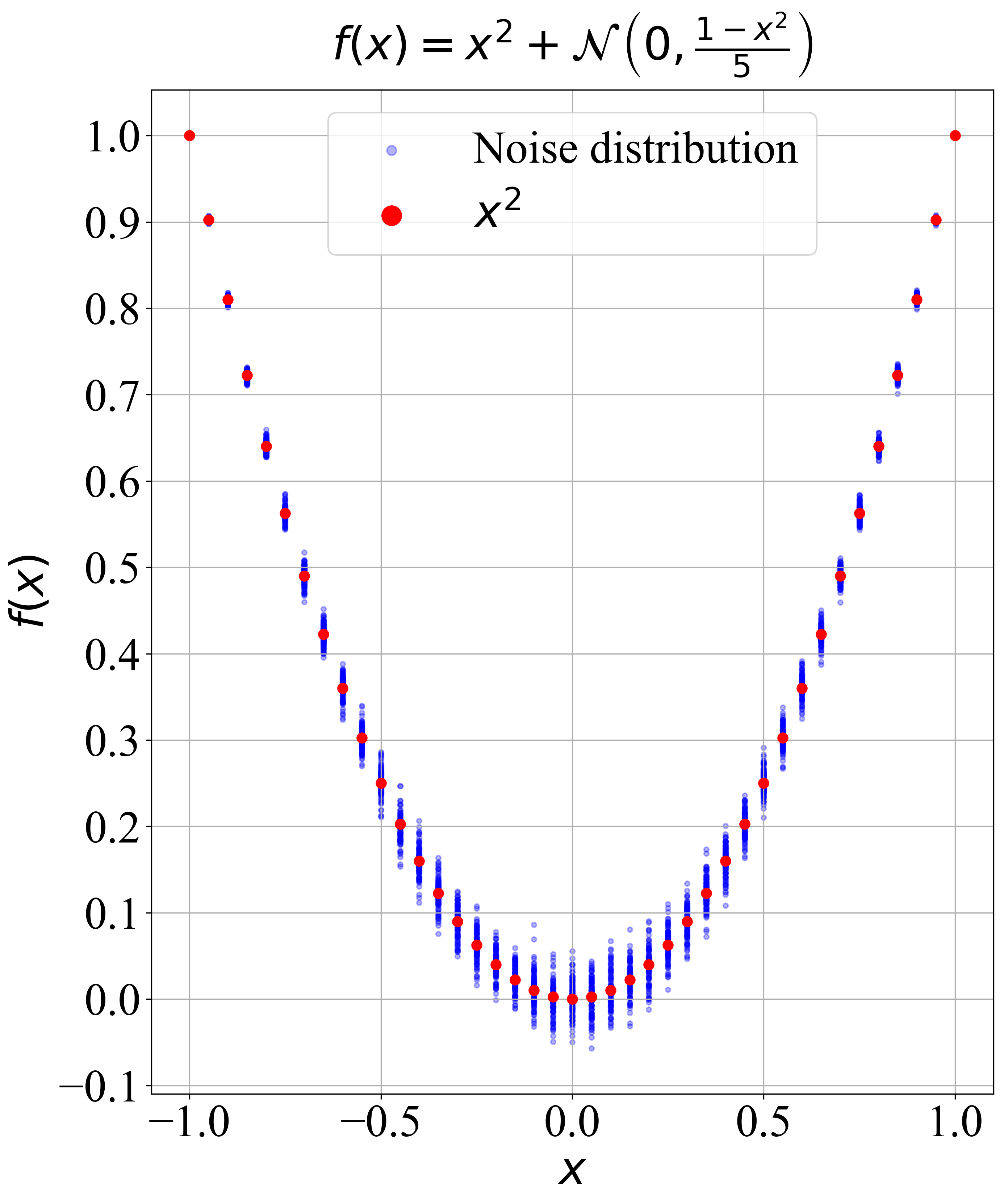}%
     \label{fig:inverse_noise}}

  \subfloat[Middle proportional noise]
    {\includegraphics[width=0.32\textwidth]{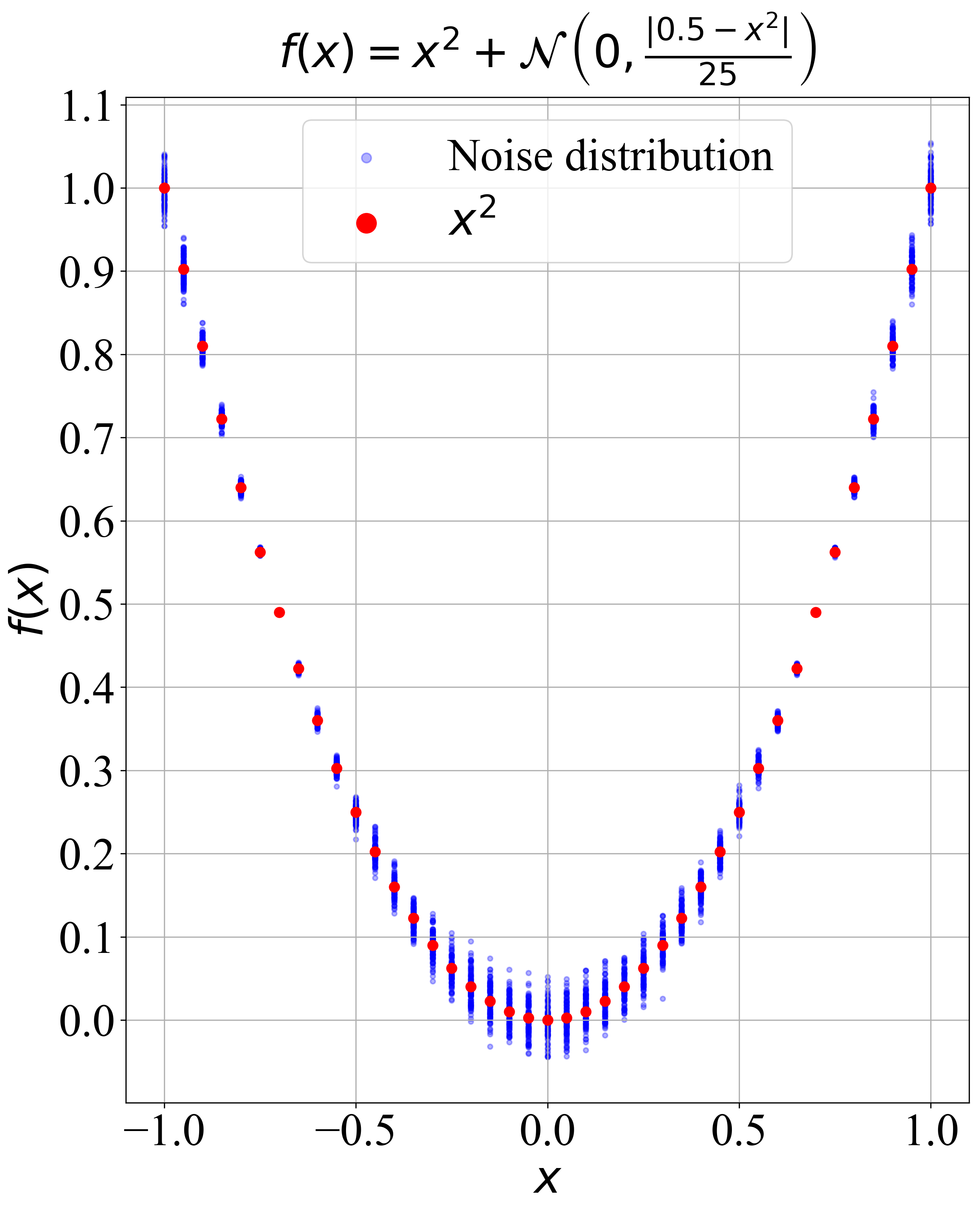}%
     \label{fig:middle_noise}}
  \hfil
  \subfloat[Sin proportional noise]
    {\includegraphics[width=0.32\textwidth]{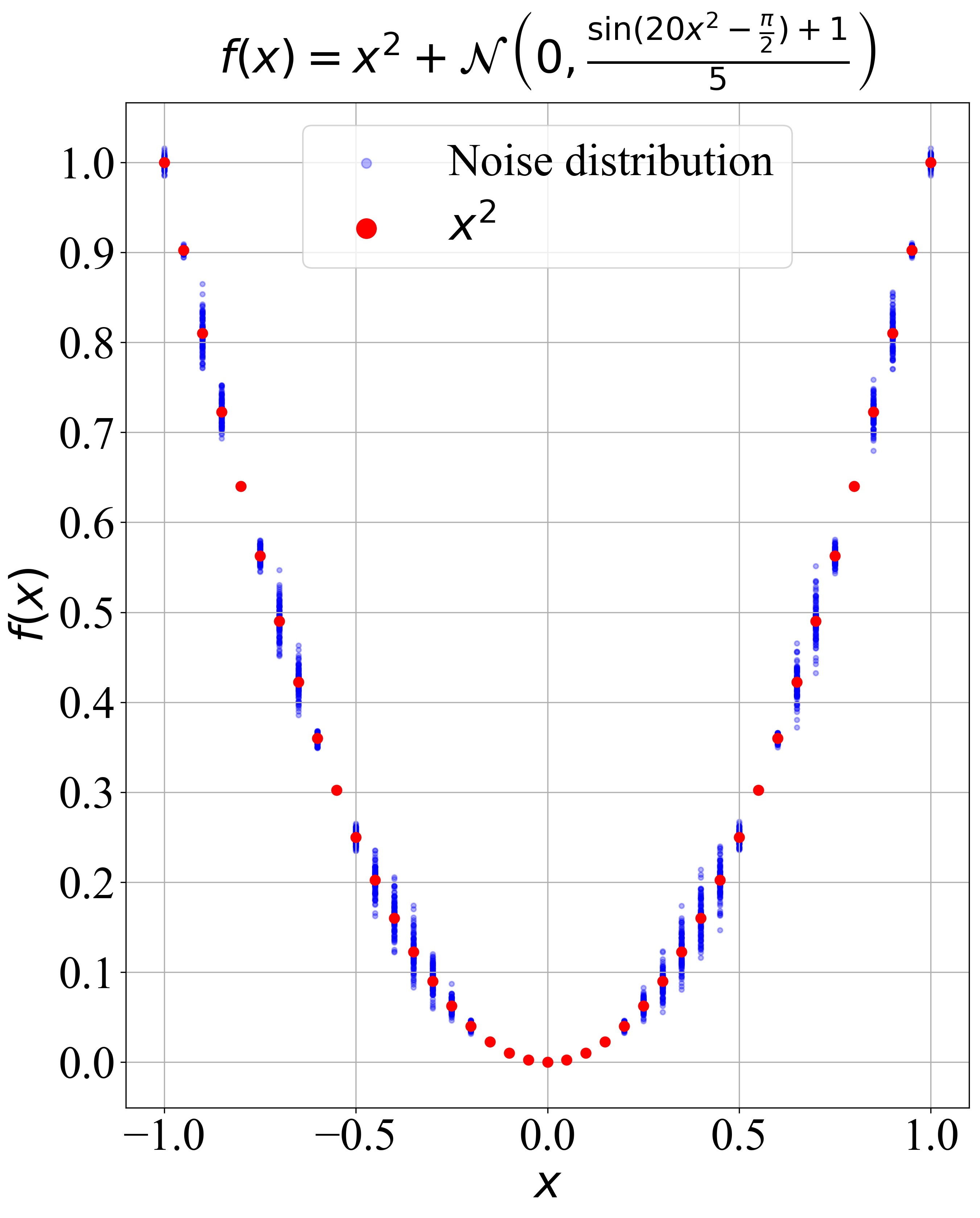}%
     \label{fig:sin_noise}}

  \caption{Illustrations of different noise profiles.}
  \label{fig:noise_structures}
\end{figure*}

To illustrate the impact of different noise profiles on optimization, 
we provide visualizations of each model. Figure~\ref{fig:constant_noise} depicts a homoscedastic noise profile, 
which represents the typical noise type employed in most prior studies within the EA domain. Figures~\ref{fig:fitness_noise}, \ref{fig:inverse_noise}, \ref{fig:middle_noise}, and \ref{fig:sin_noise} 
illustrate the four heteroscedastic noise functions used in our experiments. In all plots, the noiseless function is shown as red dots, while noisy samples are shown as blue dots.

\subsubsection{\textbf{Homoscedastic Noise}}   
In the homoscedastic noise case, every input $x$ yields noise drawn from a normal distribution with fixed standard deviation. As shown in Figure~\ref{fig:constant_noise}, the underlying function is relatively flat around the global minimum. Consequently, distinguishing which sample is better becomes more challenging in this region. In contrast, the slope is steeper farther away from the true minimum, which increases the difference in underlying function values and makes it easier to identify superior points despite the noise. This observation is consistent with the intuition of the signal-to-noise ratio (SNR) discussed in \cite{revisiting}.

\subsubsection{\textbf{Fitness Proportional Noise}}
As shown in Figure~\ref{fig:fitness_noise}, the noise variance is directly proportional to the underlying function (in the minimization case). This relationship ensures that the impact of noise, or equivalently the signal-to-noise ratio (SNR), remains constant across the search space. Consequently, the difficulty of comparing solutions is approximately uniform everywhere, 
regardless of the gradient of the objective function.

\subsubsection{\textbf{Inverse Proportional Noise}}
Under inverse-proportional noise (Figure~\ref{fig:inverse_noise}), the variance scales inversely with the underlying function.  This makes it  hard to discriminate near optimal solutions. However, it is much easier to differentiate solutions further away, particularly in steep regions. 

\subsubsection{\textbf{Middle Proportional Noise}}
Under middle-proportional noise (Figure~\ref{fig:middle_noise}), 
the variance increases as solutions move away from the midpoint between the best and worst values. This model makes it more difficult to discriminate among solutions at both extremes of the search space.

\subsubsection{\textbf{Sin Proportional Noise}}
Under sin proportional noise (Figure~\ref{fig:sin_noise}), the variance shows a cyclic pattern like a wave. This model is particularly useful for assessing whether the algorithm can allocate its evaluation budget efficiently when comparing multiple solutions with varying noise levels.

\subsection{Compared algorithms}
To demonstrate the performance of the proposed approach, we compare it against a set of peer algorithms. Several recent methods in noisy black-box optimization are selected to represent the current state-of-the-art. The original implementations of these algorithms are designed for unconstrained search and therefore do not include an explicit mechanism for handling variable bounds. However, bounded search spaces are common in practical optimisation problems, where decision variables are restricted to physically or operationally meaningful ranges. To provide a fair comparison under these conditions, we also evaluate box-constrained variants of each algorithm. The box-constrained versions were implemented by applying a simple boundary handling strategy in which any decision variable that falls outside its prescribed bounds is clipped (trimmed) to the nearest feasible bound. The several recent methods chosen for comparison are:
\begin{itemize}
    \item \textbf{OPL-CMA-ES:}  CMA-ES algorithm that re-evaluates a subset of candidate solutions to estimate the noise level and then adapts the population size accordingly \cite{OPLCMA}. Two versions of this algorithm are considered: the original variant with no bound handling (OPL-CMA-ES-UB) and a box-constrained variant (OPL-CMA-ES-B). 
    \item \textbf{LRA-CMA-ES:} CMA-ES algorithm that adjust learning rate to maintain a constant signal-to-noise ratio throughout the optimization process~\cite{LRA-CMA-ES}. Two versions of this algorithm are considered: the original variant with no bound handling (LRA-CMA-ES-UB) and a box-constrained variant (LRA-CMA-ES-B). 
    \item \textbf{RA-CMA-ES: } CMA-ES algorithm that computes two update directions, each using half of the available evaluations, and adapts the number of re-evaluations based on the estimated correlation between these two update directions \cite{RA-CMA-ES}. Two versions of this algorithm are considered: the original variant with no bound handling (RA-CMA-ES-UB) and a box-constrained variant (RA-CMA-ES-B). 
\end{itemize}

In addition, we design several additional EA variants to represent approaches previously adopted in the literature:
\begin{itemize}
    \item \textbf{Static sampling:} EA algorithm constructed as a baseline explicit averaging method, where each solution is evaluated with a fixed number of samples. A static sampling size of 30 is employed for each offspring, with ranking and offspring generation determined solely from the average of these samples. It is referred to as simple averaging~(SAVG-EA) subsequently. 
    \item \textbf{Dynamic sampling using OCBA:} EA algorithm that employs dynamic sampling based on the principles of optimal computing budget allocation~\cite{OCBA}, adjusting sample sizes proportionally to solution variances. It allocates an equivalent sampling budget of $30N$ per generation. However, it distributes samples dynamically as follows. First, the standard deviation of each solution in the population is estimated using 5 initial samples (consuming $5N$ evaluations). The remaining budget of $25N$ evaluations is then allocated to solutions in proportion to their estimated standard deviations. It is referred to as simple averaging with dynamic sampling size~(SAVGD-EA) subsequently. 
    \item \textbf{Dynamic incremental sampling:} EA algorithm that utilizes dynamic sampling with gradually increasing sample sizes. The premise is that early generations require fewer samples to obtain approximate ranks, while later generations allocate more samples to distinguish solutions that are closer to each other~\cite{incremental_sampling}.  Here, the sampling budget per generation is linearly increased as $(30+G)N$, where $G$ denotes the generation number. Thus, each solution in generation $G$ is evaluated using a fixed sample size of $30+G$.It is referred to as simple averaging with incremental sampling size~(SAVGI-EA) subsequently. 
\end{itemize}
\footnotetext{To ensure correct replication and a fair comparison, the authors of~\cite{OPLCMA} provided us with their source code. Implementations of RA-CMA-ES and LRA-CMA-ES are also publicly available from the authors of~\cite{RA-CMA-ES} at \url{https://github.com/shiralab/reevaluation-adaptation-cmaes}.}

Parameter settings used for CR-CMA-ES are provided in Table~\ref{tab:params_rc_cma}. The parameter settings of OPL-CMA-ES, RA-CMA-ES, and LRA-CMA-ES remain unchanged, except for the initialization. Specifically, the initial distribution mean $m_0$ is modified to be sampled uniformly from the range $[\textit{lower\_bound}, \textit{upper\_bound}]^D$, rather than the default $[1,5]^D$. The initial step size $\sigma$ is also adjusted to $(\textit{upper\_bound} - \textit{lower\_bound})/6$. These modifications ensure fair benchmarking, as the test problems considered in this study exhibit varying upper and lower bounds. In addition, the population sizes of RA-CMA-ES and LRA-CMA-ES are set to~20 to maintain consistency with CR-CMA-ES. 

    \begin{table}[!ht]
        \centering
        \caption{Parameter settings for CR-CMA-ES}
        \label{tab:params_rc_cma}
        \begin{tabular}{lll}
            \toprule
            Parameter & Symbol                    & Value  \\
            \midrule
            Population size & $N$                          & 20\\
            Confidence level   &       $C$       & 75\%\\
            Minimum sample size   &    $s_0$       & 3      \\
            Starting max sample size   &   $S_t$    & 30     \\
 Critical top rank solution size& $\gamma$&$0.5N$\\
            Tie threshold to update budget  & $\beta$                 & $0.5N$\\
            Global sampling budget~($B_t$) multiplier  &        $\alpha$  & 1.2      \\
            \bottomrule
        \end{tabular}
    \end{table}

Parameter settings used for CR-EA are provided in Table~\ref{tab:default-params}. The parameters for crossover and mutation for SAVG-EA, SAVGD-EA, and SAVGI-EA are kept identical to those in CR-EA. 
        \begin{table}[!ht]
        \centering
        \caption{Parameter settings for CR-EA}
        \label{tab:default-params}
        \begin{tabular}{lll}
            \toprule
            Parameter & Symbol                    & Value  \\
            \midrule
            Population size & $N$                          & 100    \\
            SBX crossover index        & $\eta_c$                        & 15     \\
            Probability of crossover  & $P_c$                  & 1.0    \\
            Polynomial mutation index         & $\eta_m$                        & 20     \\
            Probability of mutation         & $P_m$                   & 0.1    \\
            Confidence level   &       $C$       & 95\%   \\
            Minimum sample size   &    $s_0$       & 3      \\
            Starting max sample size   &   $S_t$    & 30     \\
 Critical top rank solution size& $\gamma$&1\\
            Tie threshold to update budget  & $\beta$                 & 0.2    \\
            Global sampling budget~($B_t$) multiplier  &        $\alpha$  & 2      \\
            \bottomrule
        \end{tabular}
    \end{table}

Following \cite{OPLCMA}, we set the target accuracy on the normalized function to $F(x)= 10^{-8}$. All $F(x)< 10^{-8}$ are clipped to $10^{-8}$. For each benchmark instance, 31 independent runs are conducted to observe the statistical behavior of the compared algorithms with respect to the considered performance metrics. The evaluation budget for all scenarios is set to $10{,}000D$, where $D$ denotes the number of variables. 

It is worth noting that some prior studies have employed the COCO platform with noisy BBOB problems~\cite{COCO_BBOB_Noisy} for their experiments. However, we have not adopted the platform in this study, as it tracks only the best-so-far encountered solution rather than the algorithm’s proposed best solution~\cite{COCO_BBOB_Noisy}. This limitation has been acknowledged by the developers~\cite{COCO_BBOB_Noisy}, but to our understanding the platform has not yet been updated to address it. Ideally, algorithmic performance should be evaluated based on the solution proposed at each generation, since the true (noiseless) values of candidate solutions are not available to the algorithm. In this work, we therefore implement a more realistic, representative, and fair benchmarking approach by evaluating algorithms on their suggested optimal solutions (i.e., the first-ranked solution) at every generation. An exception is made for OPL-CMA-ES, for which the reported optimal solution is the weighted average of the population \cite{OPLCMA}. 

\section{Results and Discussion}
\label{sec:results}
In this section, we summarize the results obtained from the numerical experiments. To note, since some of the algorithms does not enforce variable bounds, it was observed to occasionally produce infeasible solutions. In such cases, the best $F(x)$ reported at the end of the run corresponds to the last feasible solution suggested. If no feasible point is found during the run, it is classified as a ``fail,'' and the best objective value $F(x)$ is reported as~1. This value corresponds to a typical best solution obtained in the first generation through random sampling in the absence of noise. Each algorithm was tested over a total of 3,100 runs across all problem–noise combinations.

\subsection{Performance Profile} \label{sec:Performance Profile}
We analyze the results across all problem instances using performance profile plots~\cite{performance_profile}. The performance profile plot displays the proportion of problems each algorithm solves within a given factor of the solution obtained by the best-performing algorithm (where lower values indicate better performance). The x-axis represents the performance ratio $\tau$ (shown in logarithmic scale), indicating how many times worse an algorithm performs relative to the best one for each problem. The y-axis shows the cumulative proportion of problems $\rho(\tau)$ solved within a given ratio. A curve that rises more steeply and reaches higher values signifies a more efficient and consistent algorithm across all problem instances. 

\begin{figure}[!ht]
\centering

\begin{subfigure}{0.48\textwidth}
    \centering
    \includegraphics[width=\linewidth]{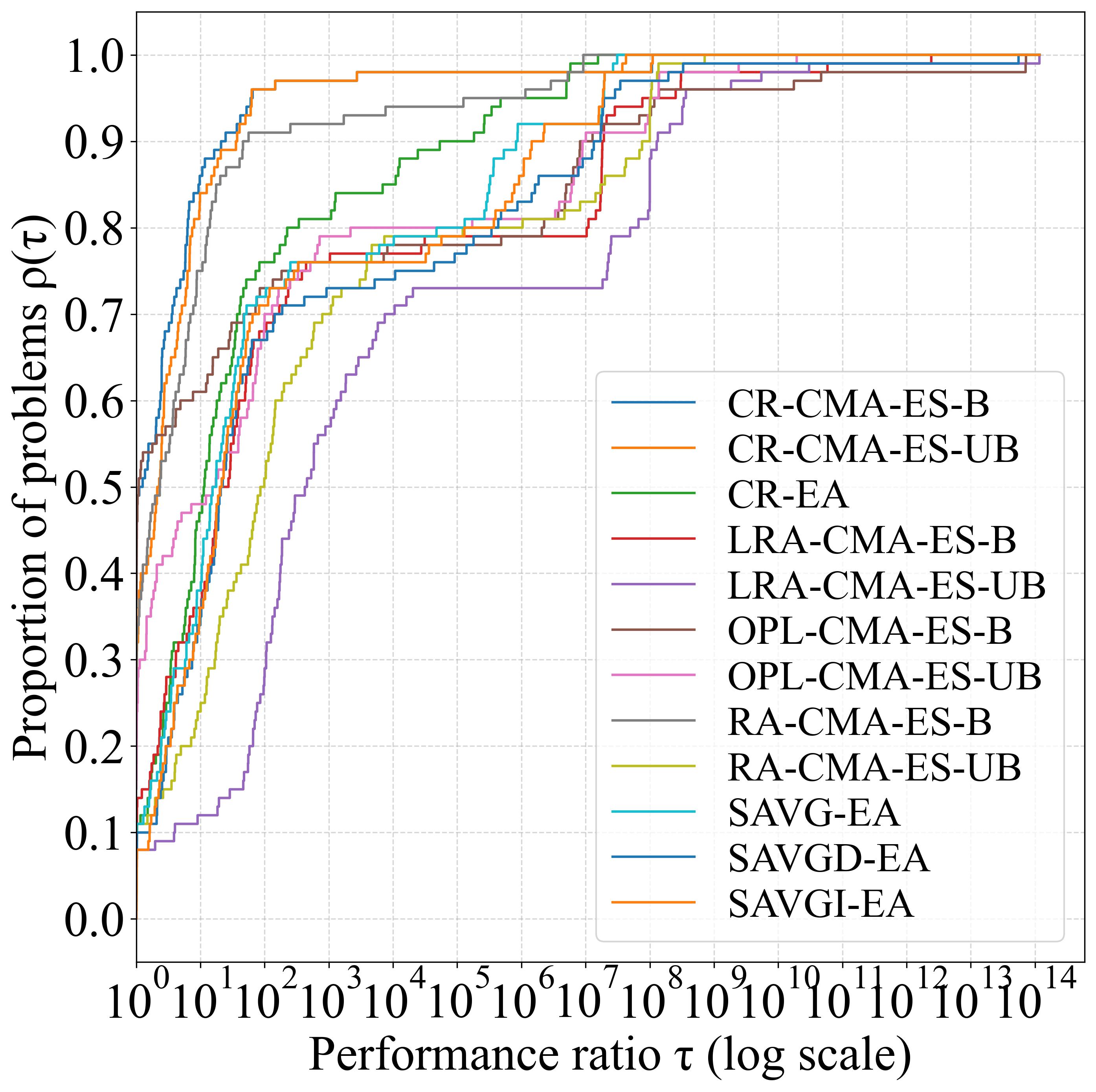}
    \caption{All Algorithms}
    \label{fig:profile_plot_all}
\end{subfigure}
\hfill
\begin{subfigure}{0.48\textwidth}
    \centering
    \includegraphics[width=\linewidth]{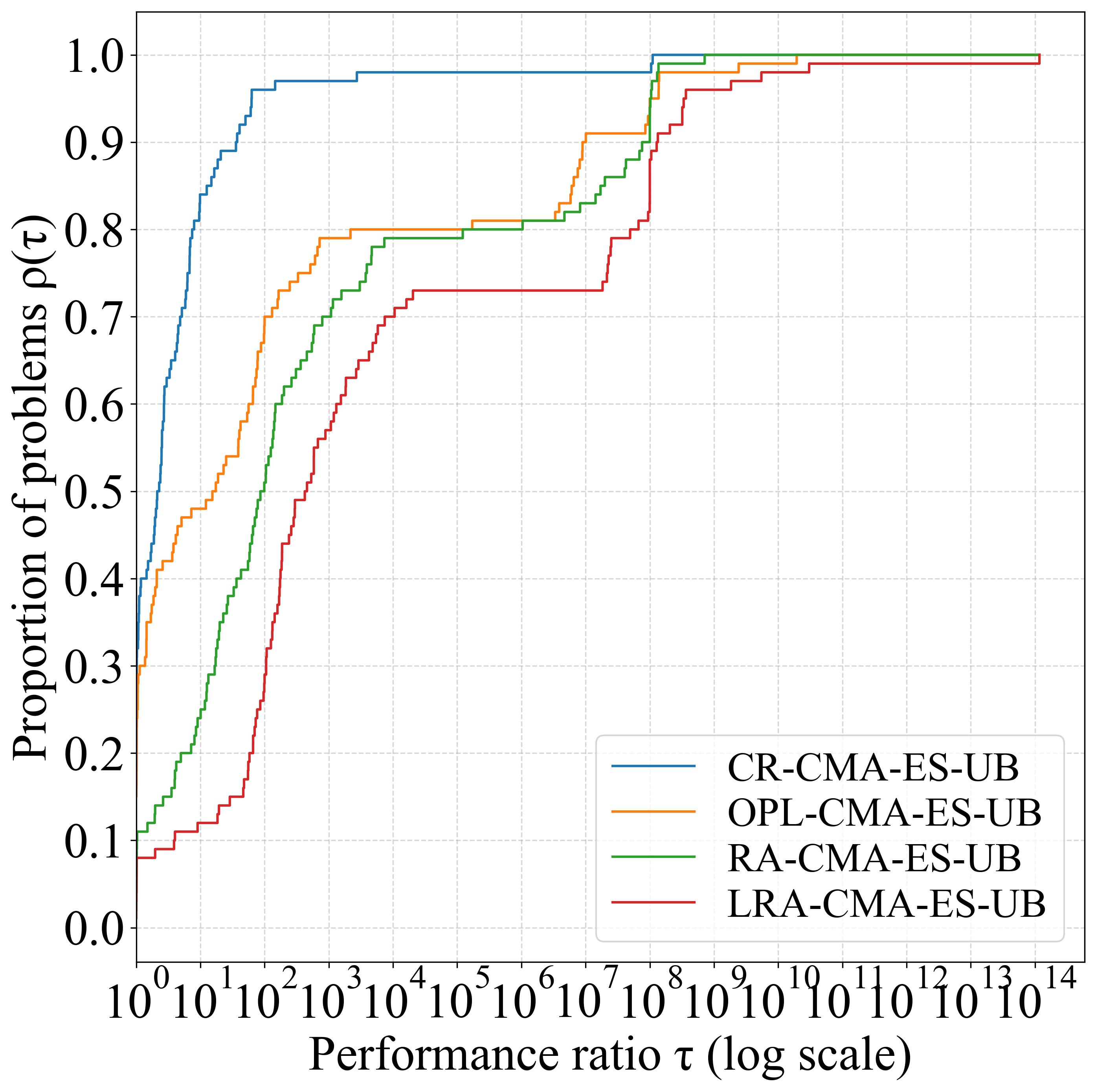}
    \caption{CMA-ES-UB algorithms}
    \label{fig:profile_plot_CMA-UB}
\end{subfigure}

\vspace{0.5em}

\begin{subfigure}{0.48\textwidth}
    \centering
    \includegraphics[width=\linewidth]{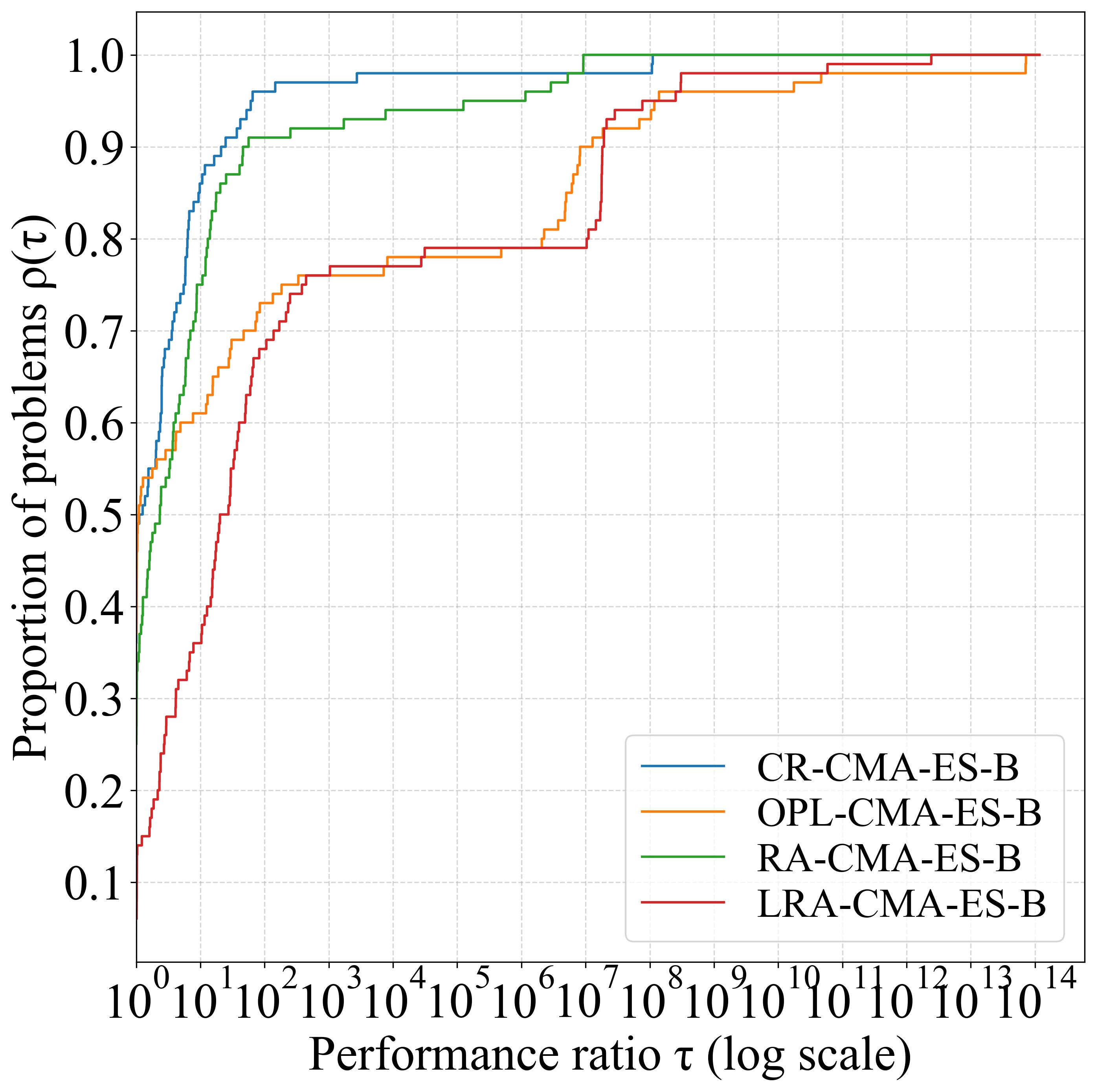}
    \caption{CMA-ES-B algorithms}
    \label{fig:profile_plot_CMA-B}
\end{subfigure}
\hfill
\begin{subfigure}{0.48\textwidth}
    \centering
    \includegraphics[width=\linewidth]{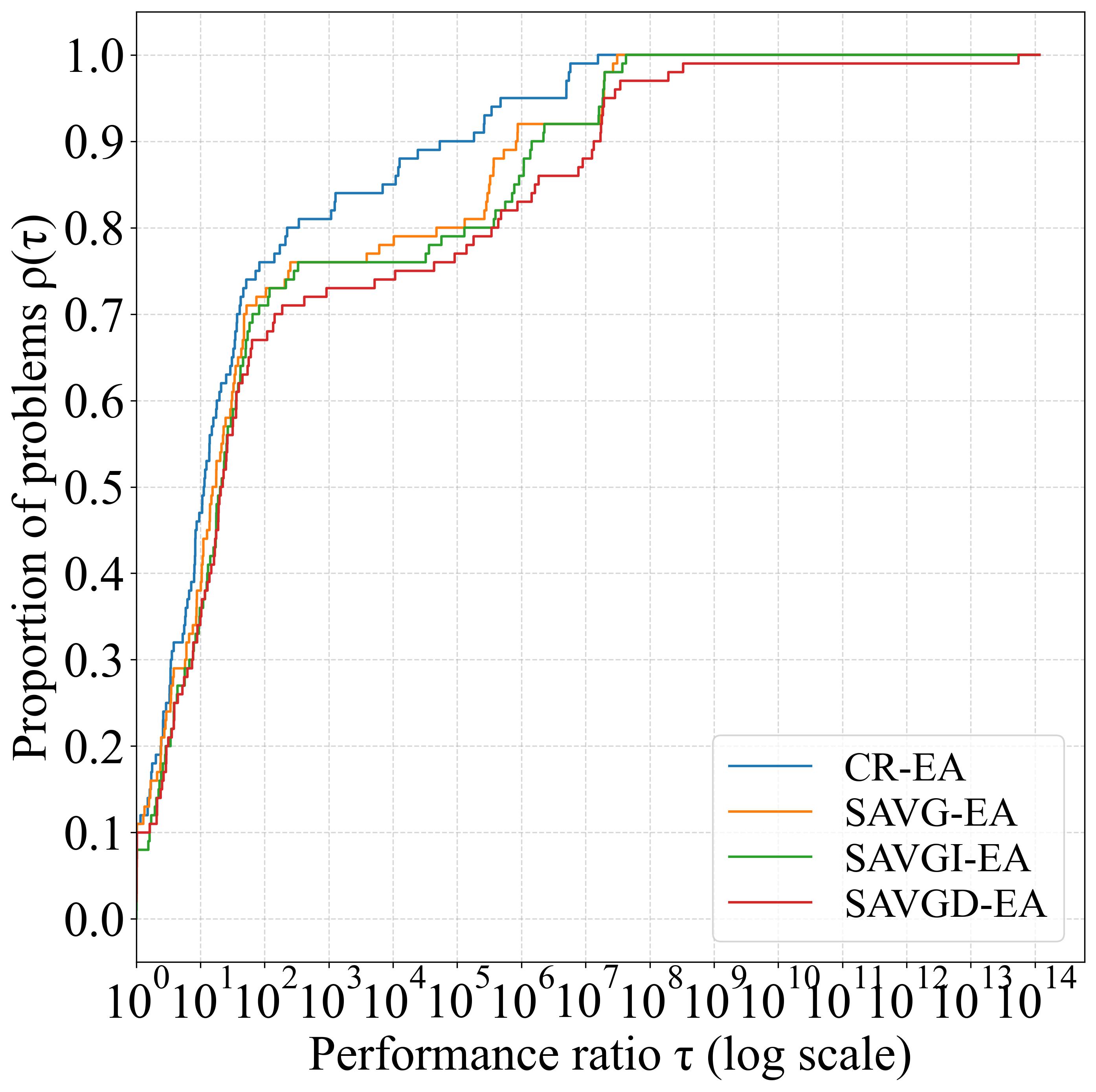}
    \caption{EA Algorithms}
    \label{fig:profile_plot_EA}
\end{subfigure}

\caption{Performance profile plots of different algorithm groups.}
\label{fig:profile_all}
\end{figure}

Four performance profiles are presented for comparison: all algorithms (figure~\ref{fig:profile_plot_all}), no bound-handling CMA-ES variants (figure~\ref{fig:profile_plot_CMA-UB}), box-constrained CMA-ES variants (figure~\ref{fig:profile_plot_CMA-B}), and genetic algorithms (figure~\ref{fig:profile_plot_EA}).

As shown in figure~\ref{fig:profile_plot_all}, CR-CMA-ES-B outperforms all other algorithms and exhibits performance comparable to its no bound-handling counterpart, CR-CMA-ES-UB. An interesting observation is that the introduction of box constraints alone leads to improved performance for LRA-CMA-ES-B, RA-CMA-ES-B, and OPL-CMA-ES-B compared to their respective no bound-handling variants. Notably, RA-CMA-ES-UB ranks as the second worst-performing algorithm, whereas RA-CMA-ES-B ranks as the third best, indicating a substantial performance improvement due to box-constraint handling.

When comparing unbounded CMA-ES variants (figure~\ref{fig:profile_plot_CMA-UB}), CR-CMA-ES-UB begins at the highest point on the left, indicating that it achieves the best result on approximately $30\%$ of the problems ($\tau = 1$). As $\tau$ increases, its performance curve consistently remains above those of all other algorithms, demonstrating that it solves a larger proportion of problems within each performance threshold. The curve’s position—higher and further to the left across most of the range—indicates not only superior performance on many problems but also greater robustness across varying difficulty levels. Even in cases where CR-CMA-ES-UB does not attain the best result, its performance remains competitive, staying within a small $\tau$ relative to other methods. A similar trend is observed for the box-constrained variants (figure~\ref{fig:profile_plot_CMA-B}). 

In figure~\ref{fig:profile_plot_CMA-B}, OPL-CMA-ES-B achieves the best result on approximately $50\%$ of the problems at $\tau = 1$, comparable to CR-CMA-ES-B. However, as $\tau$ increases, its performance curve falls below those of CR-CMA-ES-B and RA-CMA-ES-B and becomes similar to that of LRA-CMA-ES-B. This behavior indicates less consistent performance, with OPL-CMA-ES-B performing well on some problems while exhibiting weaker performance on others. In addition, it is observed that RA-CMA-ES-B overtakes CR-CMA-ES-B at $\tau = 10^7$. A closer examination of the results shows that this occurs because RA-CMA-ES-B consistently obtains better performance on all \texttt{fweierstrass} problems by exploiting the box constraint. This problem has multiple optimal solutions, many of which lie exactly on the problem boundaries. Consequently, whenever RA-CMA-ES-B generates offspring outside the feasible domain, the box constraint clips them back onto the boundary, allowing the algorithm to reach an optimal solution within only a few generations. Thus, this outlier is not due to algorithm's search abilities, but rather due to the bound constraints landing coincidentally on the problem's optima.

When the proposed method is applied to the EA framework and compared against other explicit averaging variants (figure~\ref{fig:profile_plot_EA}), CR-EA demonstrates competitive performance at small $\tau$ values and progressively outperforms the other algorithms as $\tau$ increases.

Overall, performance profile plots demonstrate that the proposed novel CR method with adaptive sampling approach yields consistently strong and consistent performance across problem settings, outperforming other methods.

\subsection{Rank Sum Tests} \label{sec: rank sum}
To assess statistical significance in performance comparisons across all problem–noise combinations, a Wilcoxon rank-sum test was conducted between:
\begin{itemize}
    \item CR-CMA-ES-UB and all other CMA-ES-UB algorithms (Table \ref{tab:cmaes_ub})
    \item CR-CMA-ES-B and all other CMA-ES-B algorithms (Table \ref{tab:cmaes_b})
    \item CR-EA and all other EA algorithms (Table \ref{tab:ea})
\end{itemize}

For each test problem, an algorithm was considered to score a \textit{win} if its performance was significantly better at the $95\%$ confidence level, a \textit{loss} if it was significantly worse, and a \textit{tie} if no significant difference was observed. 
For example in Table \ref{tab:cmaes_ub}, RA-CMA-ES-UB achieved 1 loss, 1 tie, and 8 wins against LRA-CMA-ES-UB on fitness proportional problem with noise level of $0.3$.

\begin{table}[!htbp]
\centering
\caption{Rank sum test summary of CMA-ES-UB algorithms against RC-CMA-ES-UB. Results are interpreted as \textit{RA-CMA-ES-UB's Loses~/~Ties~/~Wins}.}
\setlength{\tabcolsep}{6pt} 
\label{tab:cmaes_ub}
\begin{tabular}{c l c c c}
\toprule
Noise Level & Noise Type & LRA-CMA-ES-UB & OPL-CMA-ES-UB & RA-CMA-ES-UB \\
\midrule
0.3 & fitness proportional  & 1/1/8 & 1/6/3 & 1/1/8 \\
0.3 & homoscedastic         & 1/1/8 & 6/2/2 & 1/1/8 \\
0.3 & inverse proportional & 2/0/8 & 6/0/4 & 2/0/8 \\
0.3 & middle proportional  & 1/1/8 & 5/1/4 & 1/1/8 \\
0.3 & sin proportional      & 1/1/8 & 0/2/8 & 1/2/7 \\
1   & fitness proportional  & 1/0/9 & 0/1/9 & 1/0/9 \\
1   & homoscedastic         & 2/0/8 & 7/0/3 & 2/0/8 \\
1   & inverse proportional  & 0/2/8 & 1/2/7 & 0/2/8 \\
1   & middle proportional   & 0/1/9 & 0/1/9 & 0/1/9 \\
1   & sin proportional      & 0/2/8 & 1/0/9 & 1/1/8 \\
\midrule
\multicolumn{2}{c}{TOTAL} & 9/9/82 & 27/15/58 & 10/9/81 \\
\bottomrule
\end{tabular}
\end{table}

\begin{table}[!htbp]
\centering
\caption{Rank sum test summary of CMA-ES-B algorithms against RC-CMA-ES-B. Results are interpreted as \textit{RA-CMA-ES-B's Loses~/~Ties~/~Wins}.}
\label{tab:cmaes_b}
\setlength{\tabcolsep}{6pt} 
\begin{tabular}{c l c c c}
\toprule
Noise Level & Noise Type & LRA-CMA-ES-B & OPL-CMA-ES-B & RA-CMA-ES-B \\
\midrule
0.3 & fitness proportional  & 1/6/3  & 2/4/4  & 2/6/2 \\
0.3 & homoscedastic  & 1/0/9  & 7/1/2  & 2/3/5 \\
0.3 & inverse proportional  & 1/0/9  & 7/1/2  & 1/3/6 \\
0.3 & middle proportional   & 1/0/9  & 7/1/2  & 1/5/4 \\
0.3 & sin proportional      & 1/1/8  & 0/2/8  & 2/6/2 \\
1   & fitness proportional  & 0/1/9  & 0/1/9  & 1/2/7 \\
1   & homoscedastic  & 0/0/10 & 8/0/2  & 1/3/6 \\
1   & inverse proportional  & 1/0/9  & 6/0/4  & 1/6/3 \\
1   & middle proportional   & 0/0/10 & 0/4/6  & 1/6/3 \\
1   & sin proportional      & 1/0/9  & 0/1/9  & 1/2/7 \\
\midrule
\multicolumn{2}{c}{TOTAL} & 7/8/85 & 37/15/48 & 13/42/45 \\
\bottomrule
\end{tabular}
\end{table}

\begin{table}[!htbp]
\centering
\caption{Rank sum test summary of EA algorithms (names are abbreviated) against CR-EA. Results are interpreted as \textit{CR-EA's Loses~/~Ties~/~Wins}.}
\label{tab:ea}
\setlength{\tabcolsep}{6pt} 
\begin{tabular}{c l c c c}
\toprule
Noise Level & Noise Type & SAVG-EA & SAVGD-EA & SAVGI-EA \\
\midrule
0.3 & fitness proportional  & 0/2/8  & 0/2/8  & 0/1/9 \\
0.3 & homoscedastic  & 0/8/2  & 0/3/7  & 0/4/6 \\
0.3 & inverse proportional  & 0/9/1  & 0/4/6  & 0/4/6 \\
0.3 & middle proportional   & 0/9/1  & 0/6/4  & 0/5/5 \\
0.3 & sin proportional      & 0/2/8  & 0/1/9  & 0/1/9 \\
1   & fitness proportional  & 1/2/7  & 0/0/10 & 0/2/8 \\
1   & homoscedastic  & 0/8/2  & 0/4/6  & 0/2/8 \\
1   & inverse proportional  & 1/8/1  & 0/5/5  & 0/6/4 \\
1   & middle proportional   & 1/7/2  & 0/4/6  & 0/7/3 \\
1   & sin proportional      & 0/0/10 & 0/0/10 & 0/0/10 \\
\midrule
\multicolumn{2}{c}{TOTAL} & 3/55/42 & 0/29/71 & 0/32/68 \\
\bottomrule
\end{tabular}
\end{table}

Overall, the proposed method demonstrates consistently superior performance, as evidenced by the total number of wins across the CMA-ES-UB, CMA-ES-B, and EA categories. An interesting observation is that OPL-CMA-ES-UB and OPL-CMA-ES-B exhibit relatively stronger overall performance compared to RA-CMA-ES and LRA-CMA-ES. 

\begin{figure*}[!htbp]
  \includegraphics[width=1\textwidth]{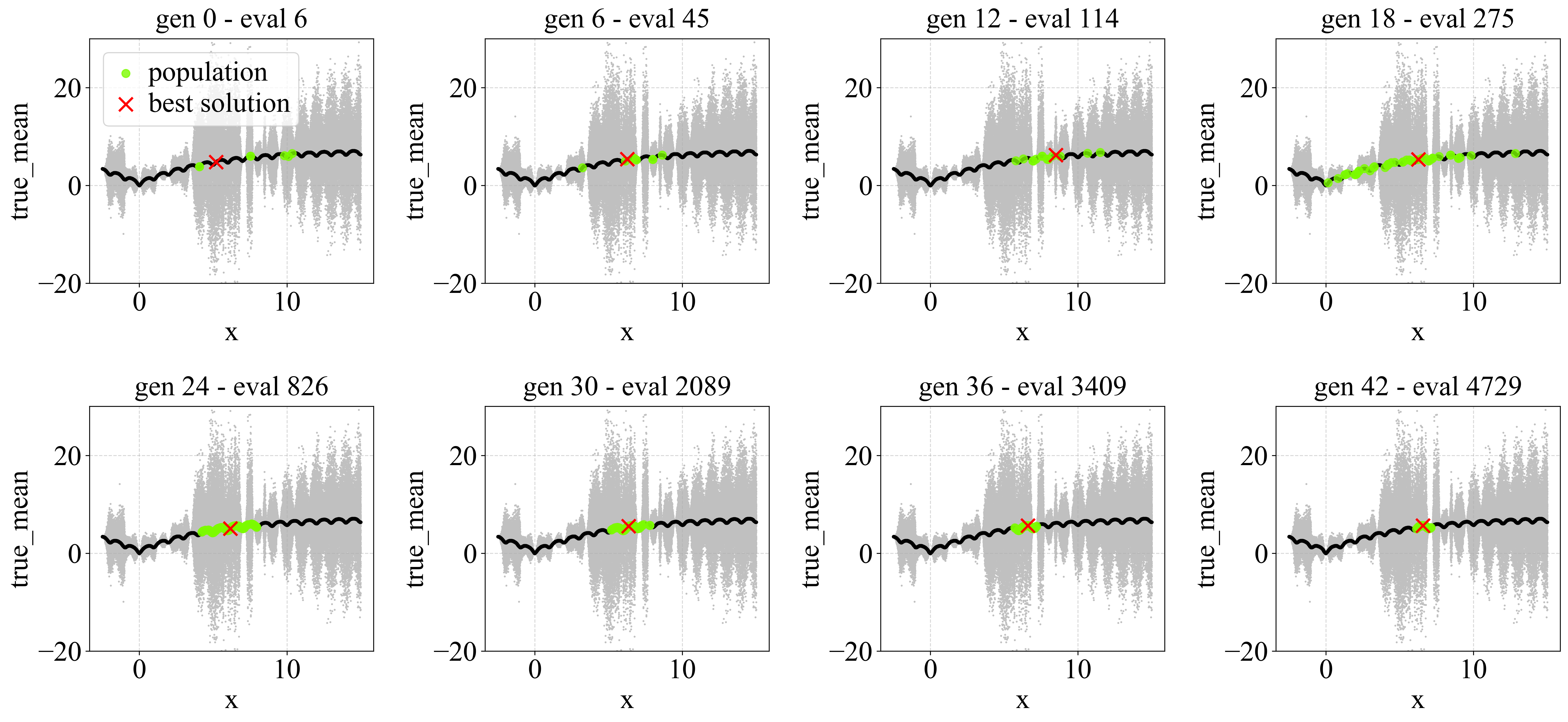}
  \caption{Movement of surviving population of OPL-CMA-ES-UB over a sample run for Ackley problem, while subject to sin proportional noise with $c=2$.}
  \label{fig:behaviour_oplcma}
\end{figure*}

\begin{figure*}[!htbp]
  \includegraphics[width=1\textwidth]{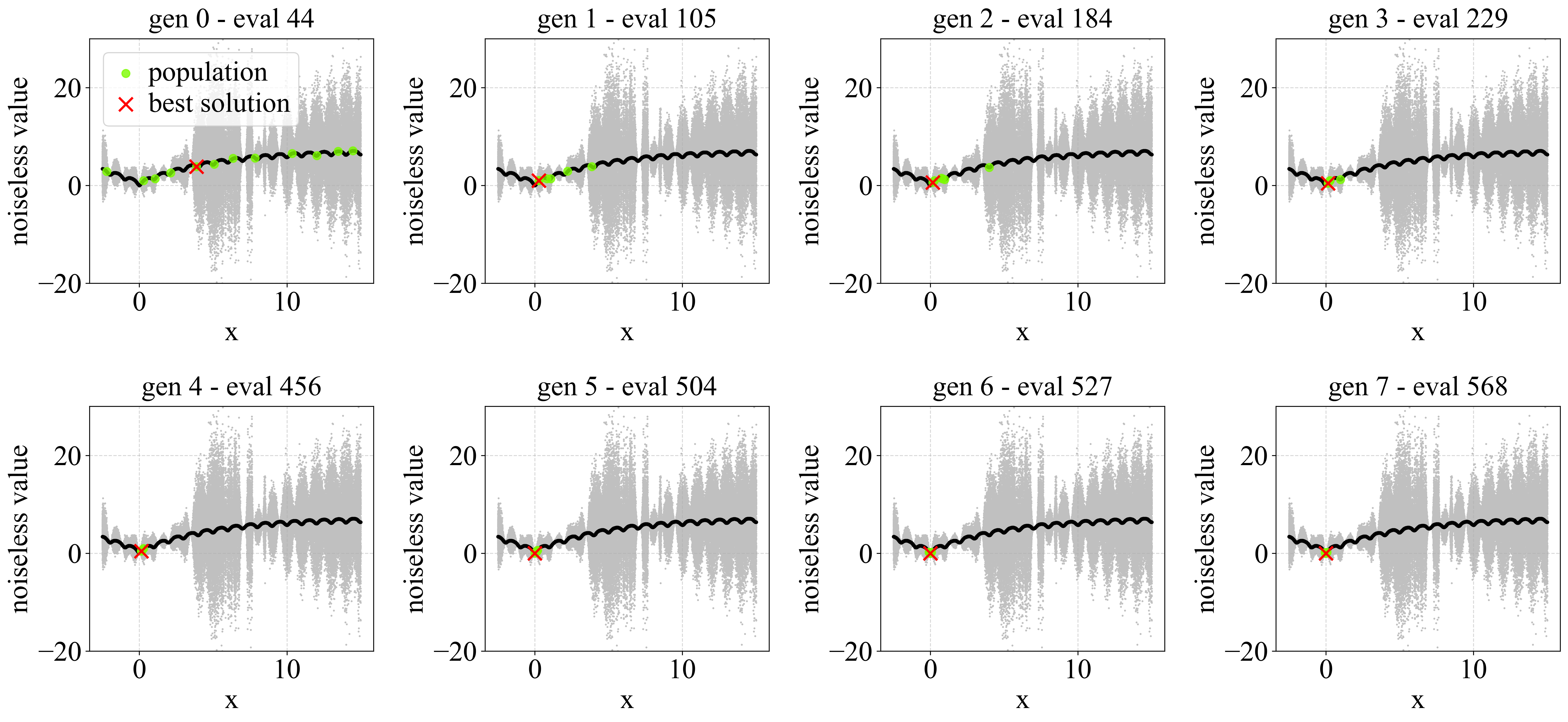}
  \caption{Movement of surviving population of CR-EA over a sample run for Ackley problem, while subject to sin proportional noise with $c=2$.}
  \label{fig:behaviour_rico}
\end{figure*}

A closer examination of the results reveals that OPL-CMA-ES achieves strong performance on certain problem instances while performing poorly on others. For example, OPL-CMA-ES-B performs particularly well under homoscedastic noise at noise level~1, but performs poorly under sinusoidal proportional noise at the same noise level when compared to RA-CMA-ES-B. This variability in performance is consistent with the trends observed in previous section (figure~\ref{fig:profile_plot_CMA-UB} and figure~\ref{fig:profile_plot_CMA-B}).

OPL-CMA-ES’s high performance on certain classes of problems stem from its ranking strategy, where the population is ordered based on noisy evaluations and the best solution is reported as a weighted average of the population. While this approach can be effective under low-noise conditions or when the gradient is steep, it may be misled when noisy evaluations appear superior to surrounding solutions. In contrast, the novel CR method with adaptive sampling adopts a more conservative strategy by allocating additional evaluations to promising solutions, thereby reducing the likelihood of being deceived by noise. 

To illustrate this difference, we conduct an experiment on 1 dimensional Ackley problem with bounds $[-2.5,15]$, subject to sin proportional noise with $c=2$. Figures~\ref{fig:behaviour_oplcma} and~\ref{fig:behaviour_rico} visualize the movement of the surviving populations of OPL-CMA-ES-UB and CR-EA, respectively. As illustrated,  OPL-CMA-ES-UB (Figure~\ref{fig:behaviour_oplcma}) is misguided toward a region of high noise, whereas CR-EA (Figure~\ref{fig:behaviour_rico}) successfully converges to the location of global minimum.

Additional illustration of RA-CMA-ES-UB and LRA-CMA-ES-UB movement are provided in Section~I of \texttt{supplementary material.pdf}.

\subsection{Convergence Trends}

\begin{figure*}[!htbp]
  \centering

  \subfloat[CMA-ES-UB (1)]{%
    \includegraphics[width=0.32\textwidth]{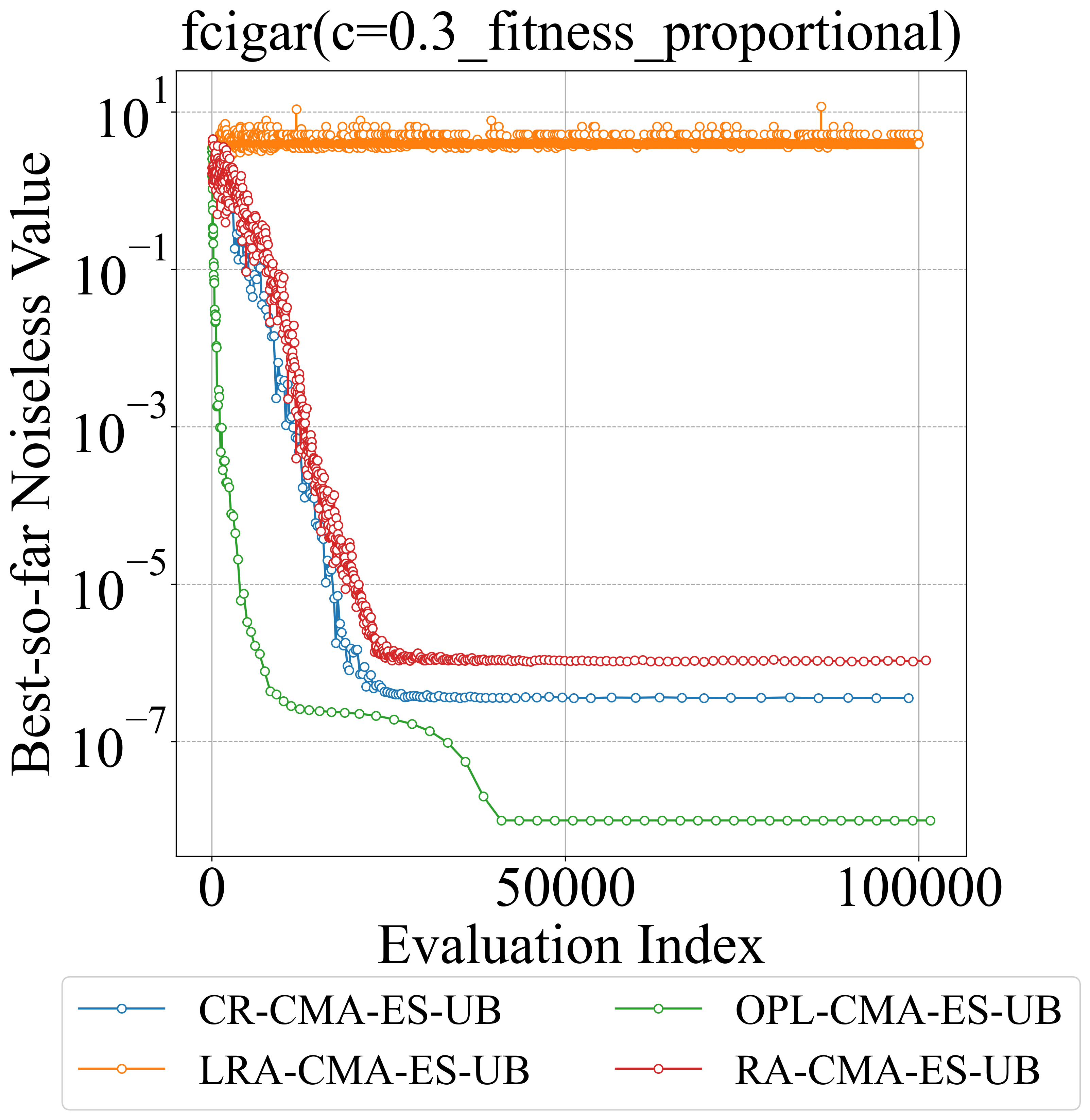}%
    \label{fig:CMA-ES-UB (1)}%
  }
  \hfil
  \subfloat[CMA-ES-UB (2)]{%
    \includegraphics[width=0.32\textwidth]{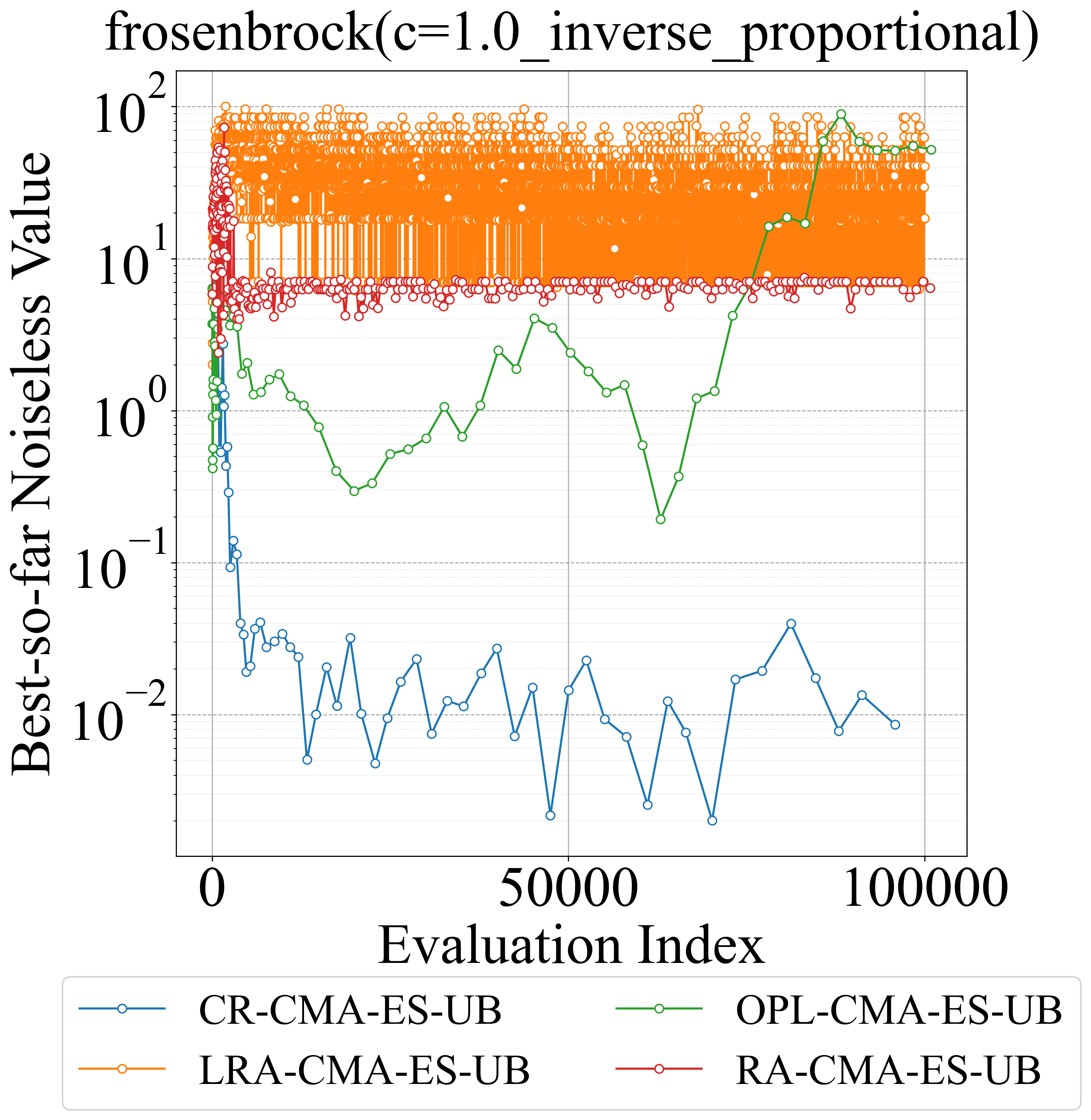}%
    \label{fig:CMA-ES-UB (2)}%
  }
  \hfil
  \subfloat[CMA-ES-UB (3)]{%
    \includegraphics[width=0.32\textwidth]{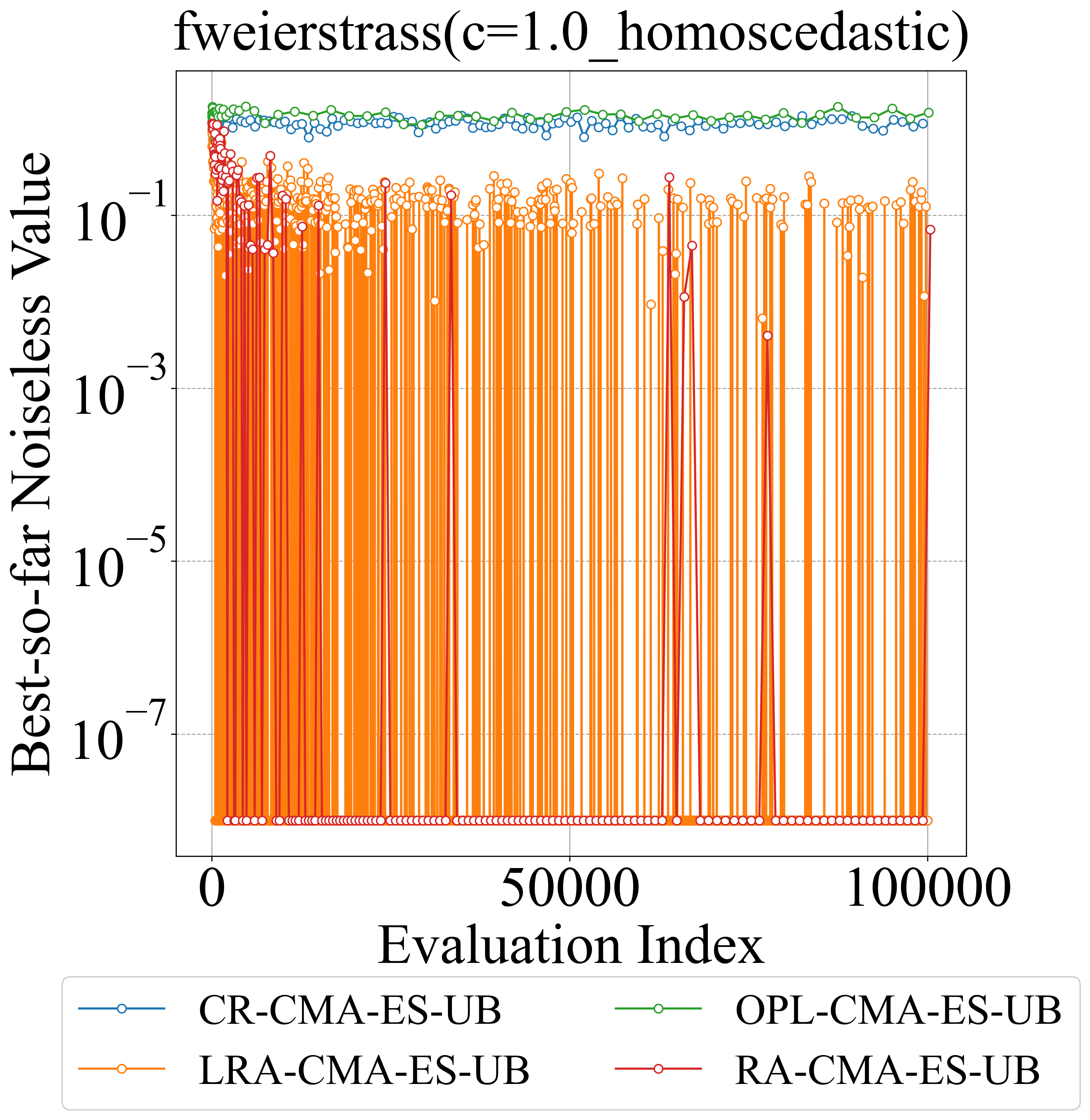}%
    \label{fig:CMA-ES-UB (3)}%
  }

  \subfloat[CMA-ES-B (1)]{%
    \includegraphics[width=0.32\textwidth]{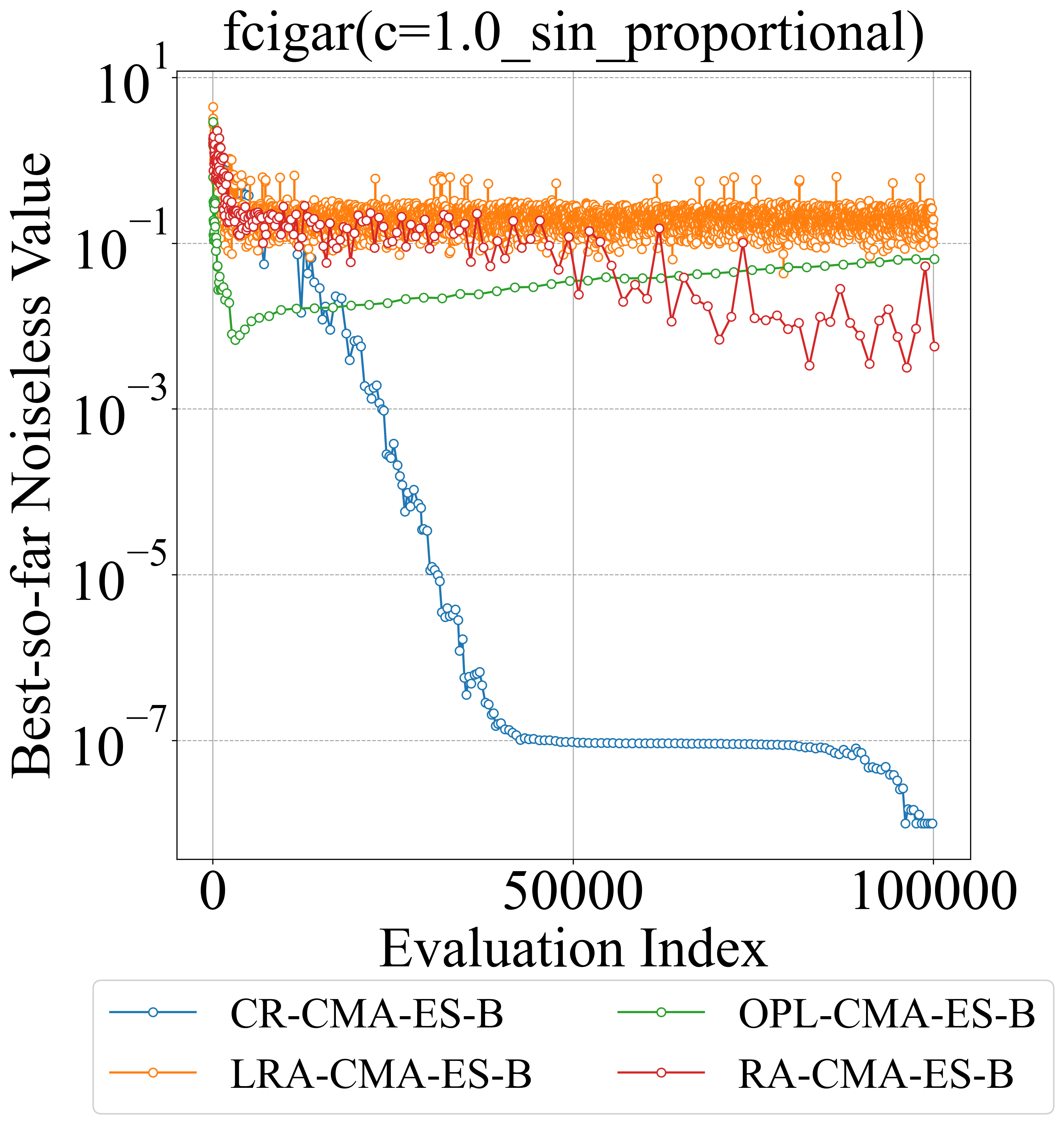}%
    \label{fig:CMA-ES-B (1)}%
  }
  \hfil
  \subfloat[CMA-ES-B (2)]{%
    \includegraphics[width=0.32\textwidth]{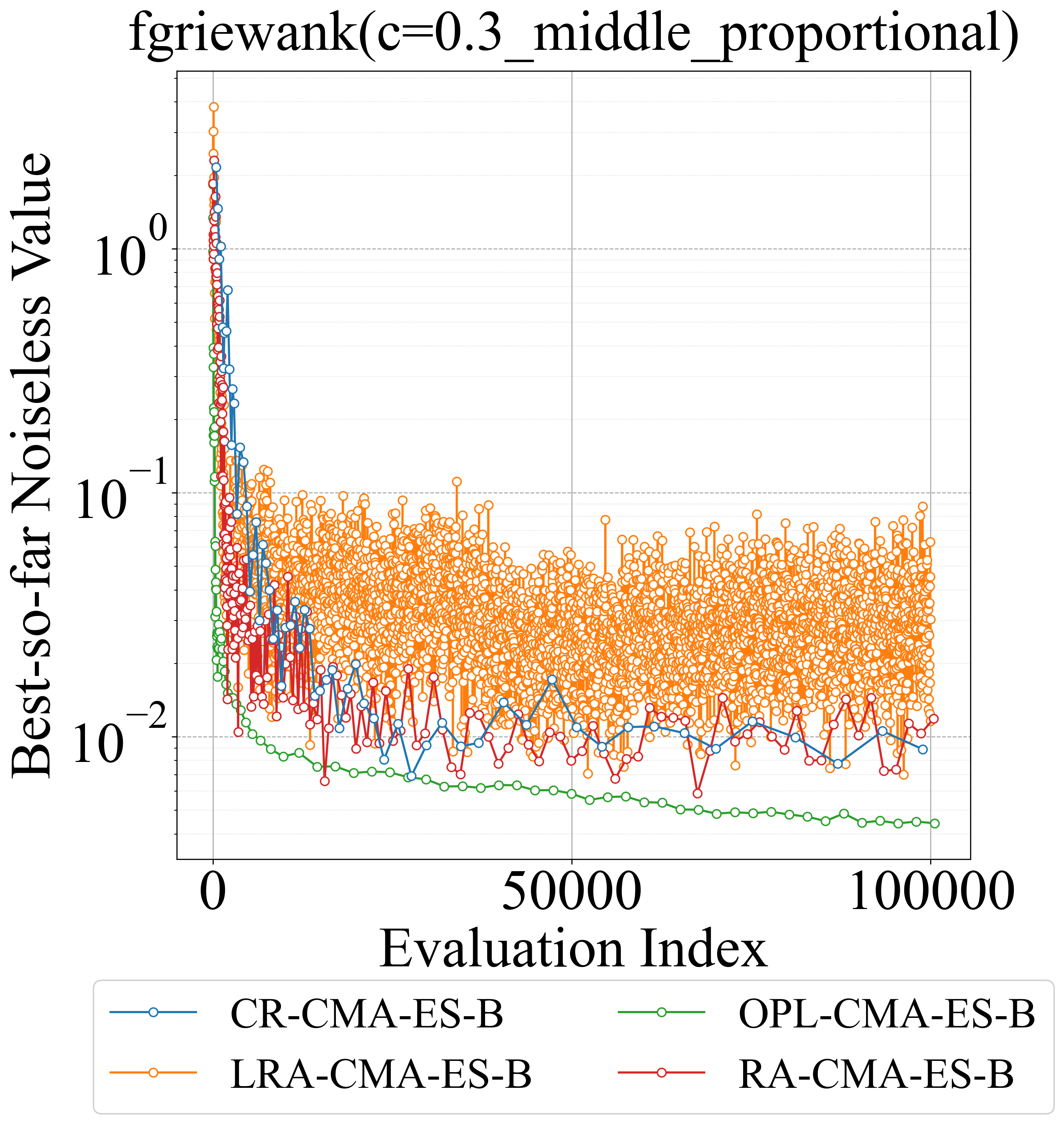}%
    \label{fig:CMA-ES-B (2)}%
  }
  \hfil
  \subfloat[CMA-ES-B (3)]{%
    \includegraphics[width=0.32\textwidth]{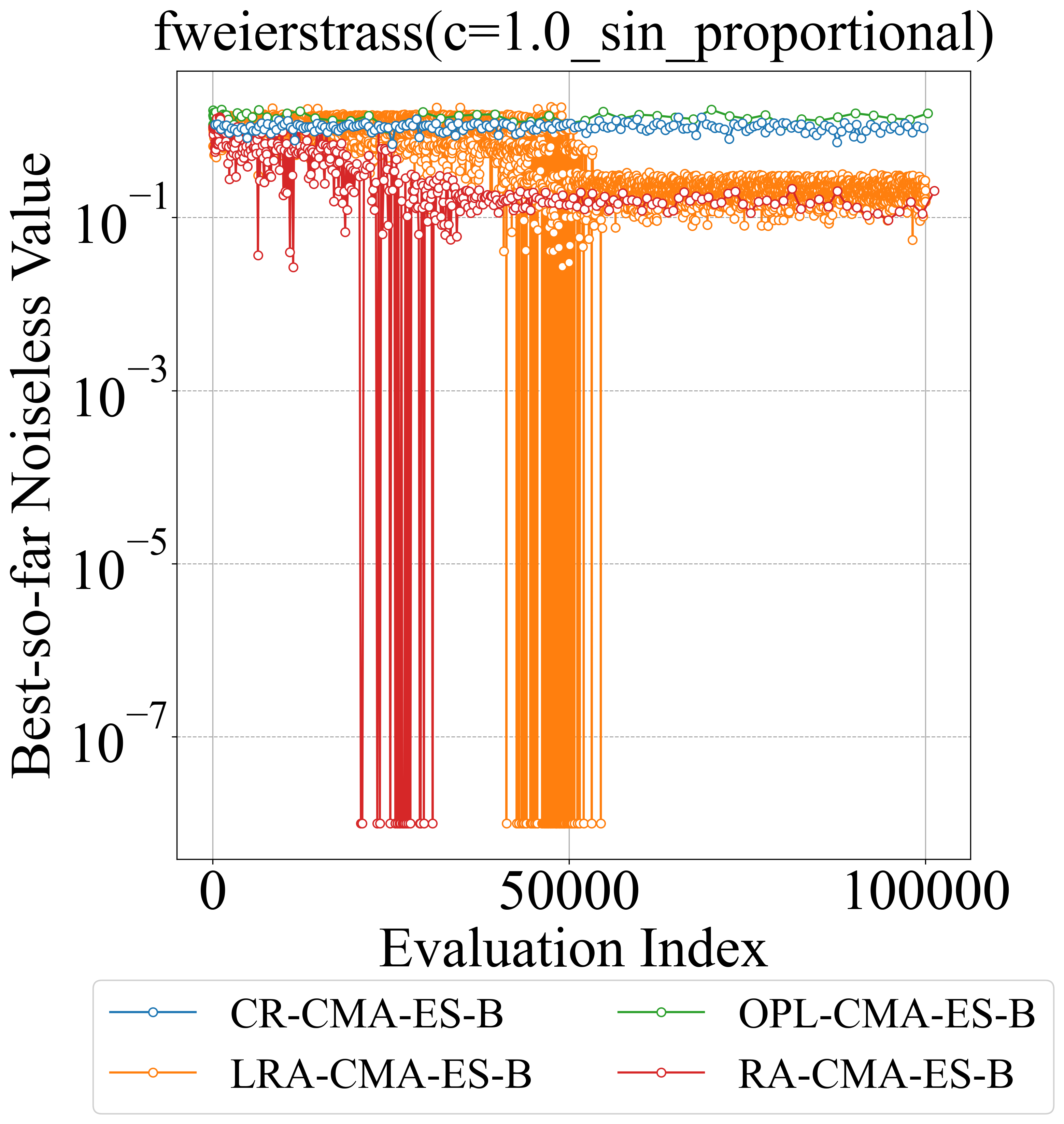}%
    \label{fig:CMA-ES-B (3)}%
  }

  \subfloat[EA (1)]{%
    \includegraphics[width=0.32\textwidth]{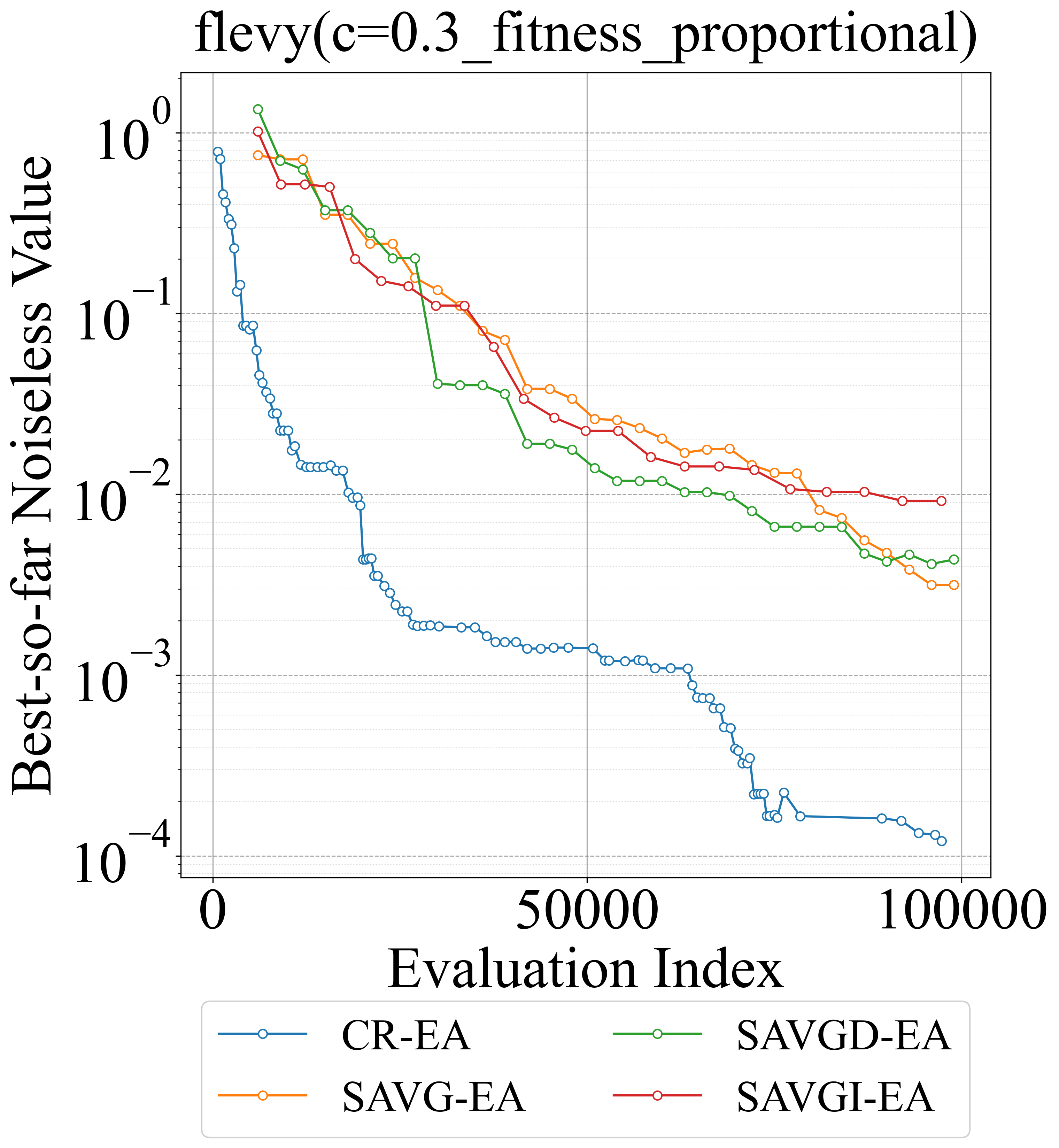}%
    \label{fig:EA (1)}%
  }
  \hfil
  \subfloat[EA (2)]{%
    \includegraphics[width=0.32\textwidth]{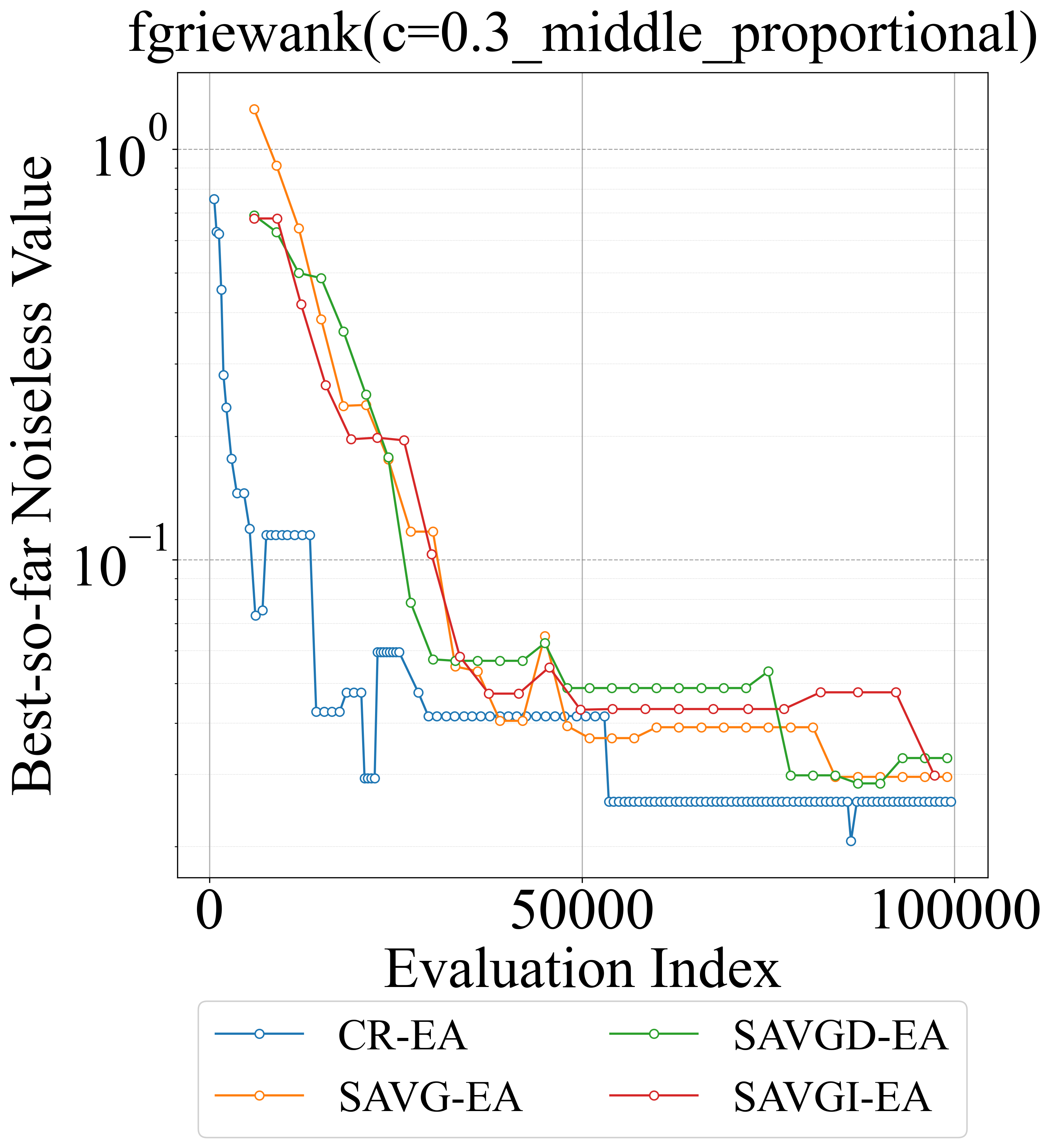}%
    \label{fig:EA (2)}%
  }
  \hfil
  \subfloat[EA (3)]{%
    \includegraphics[width=0.32\textwidth]{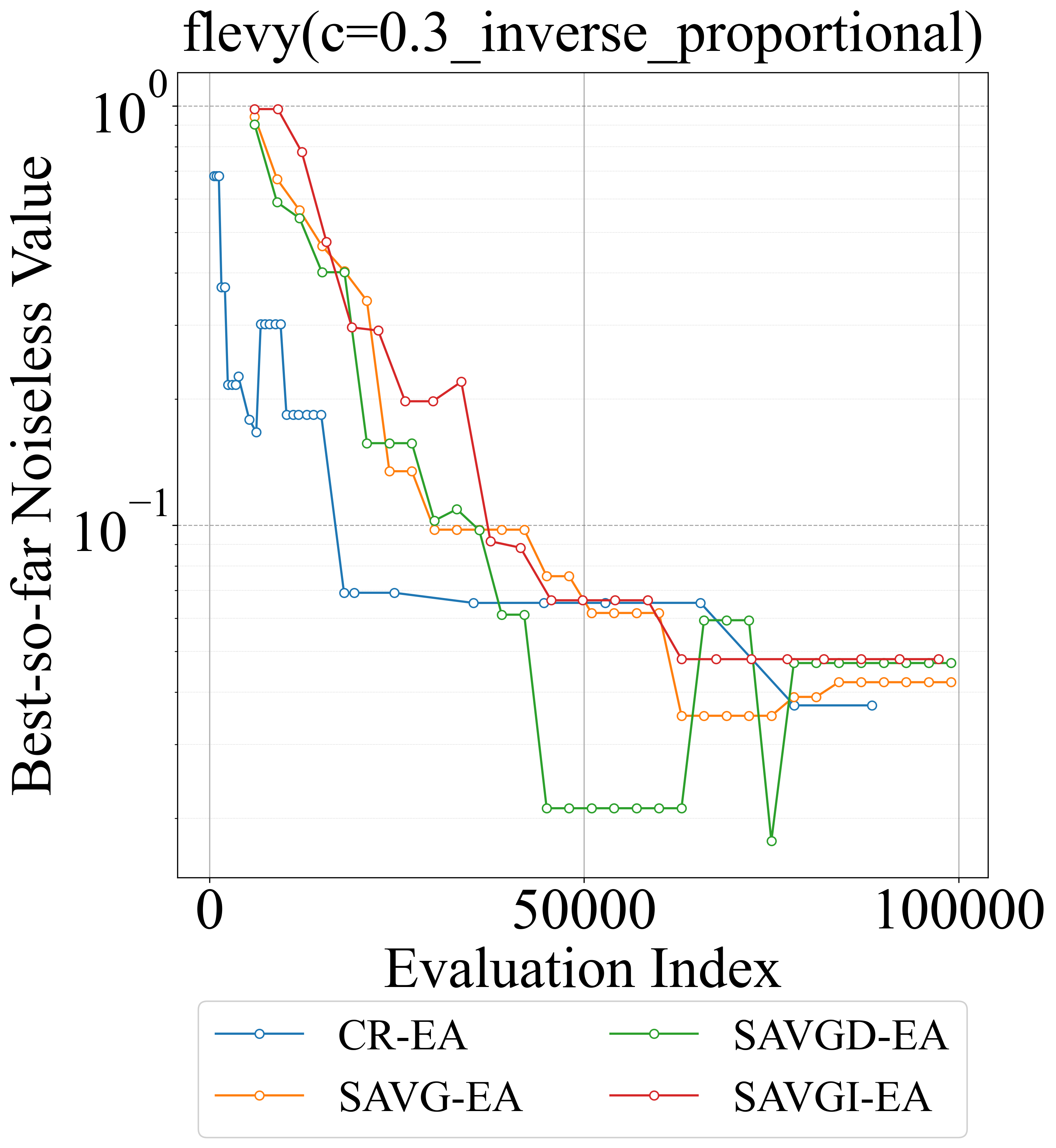}%
    \label{fig:EA (3)}%
  }

  \caption{Representative convergence plots for different scenarios and algorithms. For each instance, the best noiseless value over the median run is shown.}
  \label{fig:representative_conv_plots}
\end{figure*}

To observe the convergence trends for each problem–noise combination, we first identify the median run by ranking all independent runs according to the noiseless $F(x)$ of the last proposed best solution before exceeding the evaluation budget. The convergence plot then displays the algorithm’s proposed best noiseless $F(x)$ at each generation of that median run.

The complete set of convergence plots are provided in the \texttt{supplementary material.zip} file (folder: \textit{Convergence Plots}). Overall, the results indicate that CR-CMA-ES-UB, CR-CMA-ES-B, and CR-EA converge faster than their respective peer algorithms in the same class, consistent with the findings previously discussed in Sections~\ref{sec:Performance Profile} and~\ref{sec: rank sum}.

To highlight the range of behaviors exhibited by each algorithm, figure~\ref{fig:representative_conv_plots} presents a collection of representative convergence plots drawn from different scenarios: CMA-ES-UB (Figs.~\ref{fig:CMA-ES-UB (1)}, \ref{fig:CMA-ES-UB (2)}, and \ref{fig:CMA-ES-UB (3)}), CMA-ES-B (Figs.~\ref{fig:CMA-ES-B (1)}, \ref{fig:CMA-ES-B (2)}, and \ref{fig:CMA-ES-B (3)}), and EA (Figs.~\ref{fig:EA (1)}, \ref{fig:EA (2)}, and \ref{fig:EA (3)}).

As discussed in previous section, OPL-CMA-ES can outperform other methods on certain problem classes while performing poorly on others. One contributing factor is its susceptibility to being misled by noise (see figure~\ref{fig:behaviour_oplcma}). Figures~\ref{fig:CMA-ES-UB (1)} and~\ref{fig:CMA-ES-B (2)} illustrate cases in which OPL-CMA-ES exhibits superior performance, whereas Figs.~\ref{fig:CMA-ES-UB (2)} and~\ref{fig:CMA-ES-B (1)} demonstrate scenarios in which it is deceived by noise and its performance deteriorates as the search progresses.

In Figs.~\ref{fig:CMA-ES-UB (3)} and~\ref{fig:CMA-ES-B (3)}, RA-CMA-ES and LRA-CMA-ES exhibit an additional noteworthy behavior: the proposed best solution along generations is highly volatile, fluctuating between approximately $10^{-8}$ and $10^{-1}$. This variability highlights instability in performance, although RA-CMA-ES appears more stable than LRA-CMA-ES. Notably, in figure~\ref{fig:CMA-ES-UB (3)}, although RA-CMA-ES-UB successfully identifies the optimal solution, it fails to retain it and ultimately reports a suboptimal solution in the final generation.

Finally, figure~\ref{fig:EA (1)} demonstrates cases in which CR-EA clearly outperforms other EA variants, while Figs.~\ref{fig:EA (2)} and~\ref{fig:EA (3)} show scenarios where performance is competitive.

\subsection{Rank Correlation} 
\label{sec:true_rank}

\begin{figure}[!htbp]
  \subfloat[CR-EA (1)]
    {\includegraphics[width=0.48\textwidth]{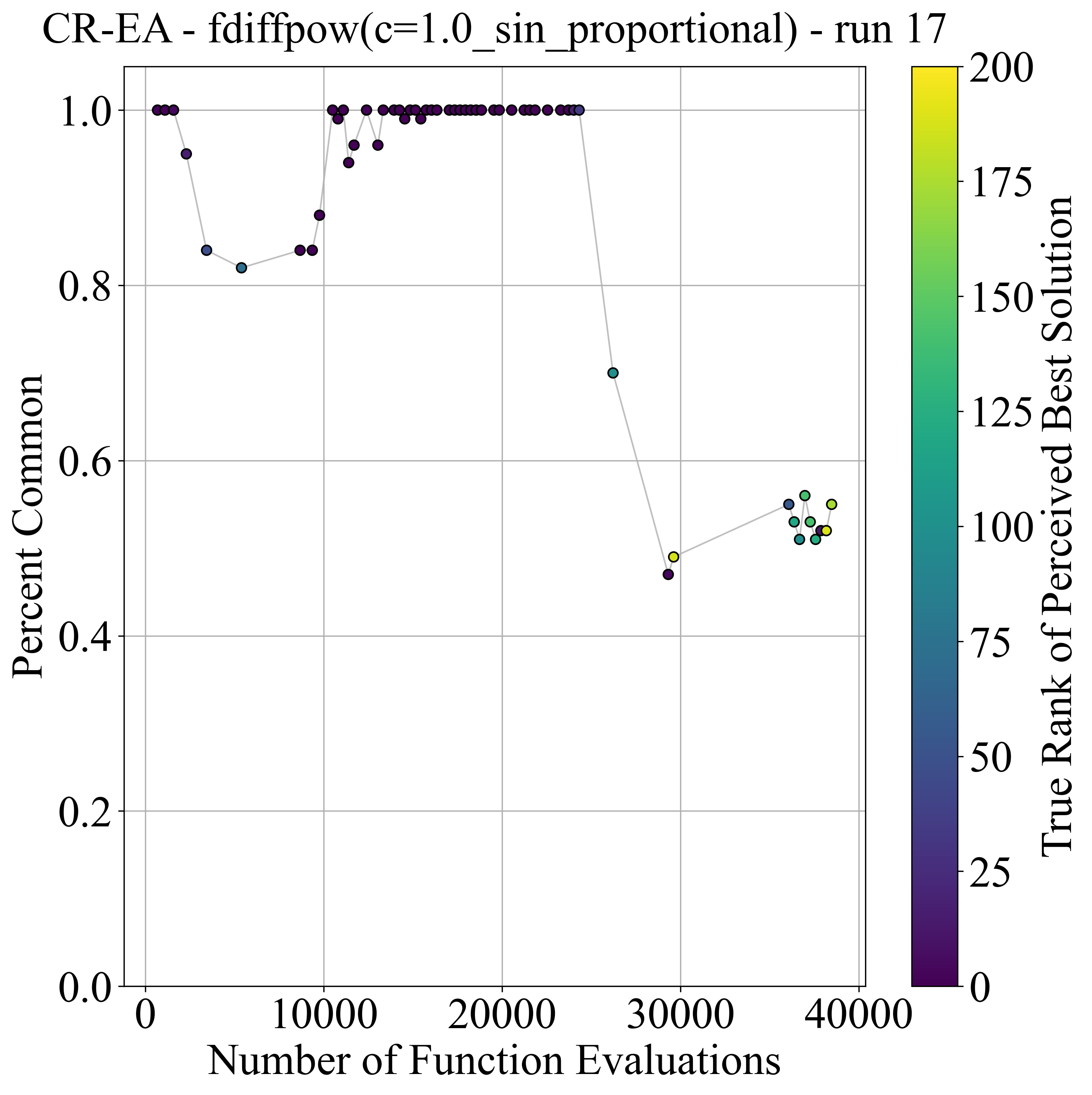}%
     \label{fig:ranking_CR-EA_1}}
  \hfil
  \subfloat[CR-EA (2)]
    {\includegraphics[width=0.48\textwidth]{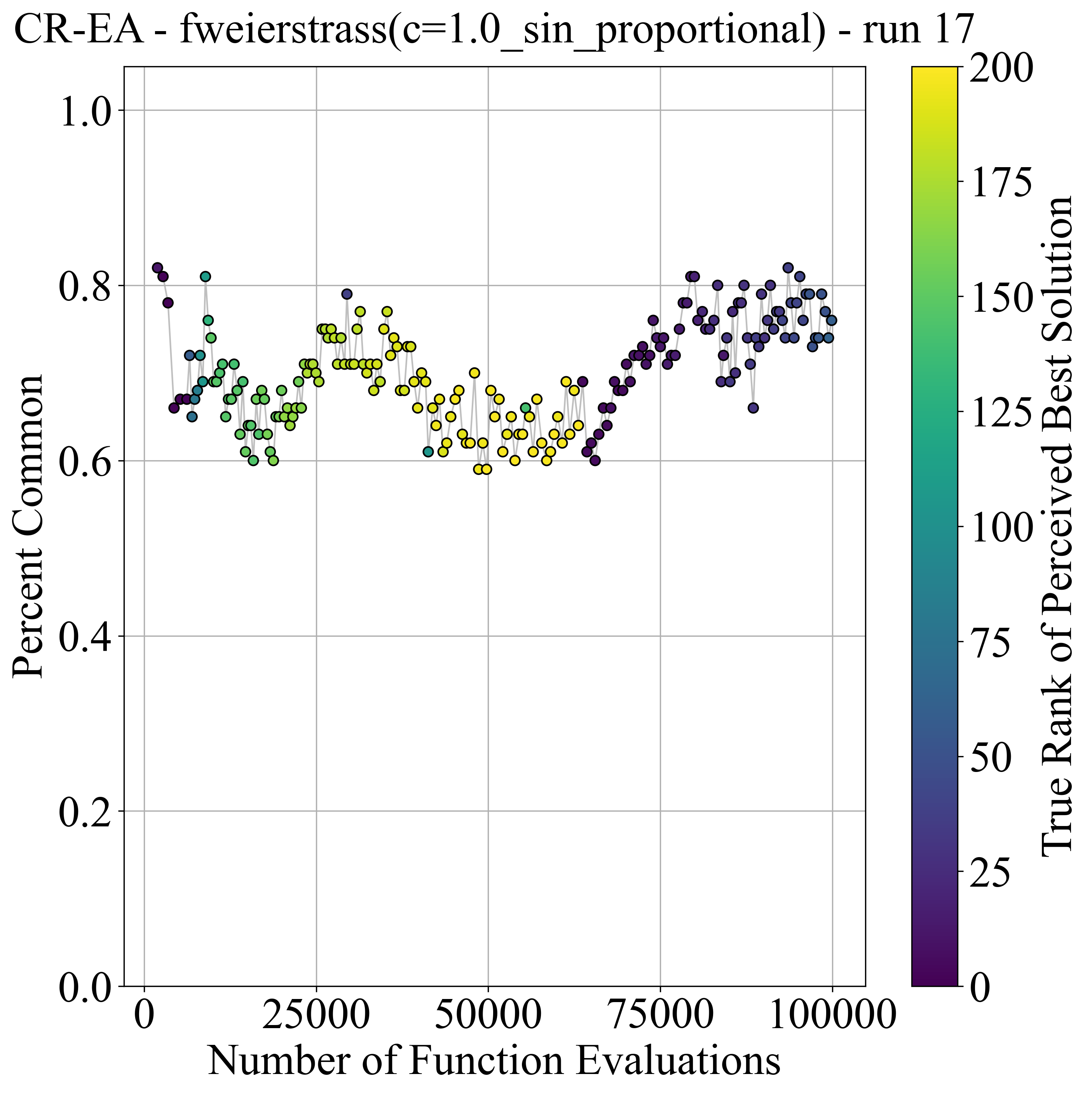}%
     \label{fig:ranking_CR-EA_2}}

  \caption{True rank plot of CR-EA}
  \label{fig:ranking_CR-EA}
\end{figure}

As CR-EA employs elitism to guide its search strategy, correctly identifying the surviving population is crucial. To further demonstrate the effectiveness of the proposed ranking method within an EA framework, we present the True-Rank plot (figure~\ref{fig:ranking_CR-EA}). This plot tracks how accurately the algorithm identifies the first-ranked solution and the surviving population relative to the true (noiseless) ranking at each generation. The color bar indicates the true rank of the algorithm's selected first-rank solution in each generation, while the $y$-axis represents the proportion of elite solutions that are carried forward.

Figures~\ref{fig:ranking_CR-EA_1} and~\ref{fig:ranking_CR-EA_2} provide representative examples illustrating the behavior of CR-EA when combined with the proposed CR method with adaptive sampling strategy. The complete set of True-Rank plots for CR-EA is included in the \texttt{supplementary material.zip} file (folder: \textit{True Rank Plots}).

Figure~\ref{fig:ranking_CR-EA_1} illustrates a setting in which CR-EA maintains highly consistent performance, accurately identifying both the first-ranked solution and the surviving population in most generations. In generations where the $y$-axis reaches $1.0$, the problem is effectively reduced to a deterministic setting. However, performance degradation is observed in later generations. This occurs because the algorithm has largely converged, resulting in a reduced signal-to-noise ratio. Consequently, CR-EA can no longer reliably distinguish small differences among solutions and reaches its statistical limit. To further improve performance beyond this stage, the confidence-level parameter must be increased. Note that the $x$-axis in figure~\ref{fig:ranking_CR-EA_2} ends at 40{,}000 function evaluations, as subsequent generations exceed the evaluation budget and are therefore omitted.

Figure~\ref{fig:ranking_CR-EA_2} illustrates a scenario in which CR-EA encounters difficulties. In this case, the algorithm retains approximately $60\%$ to $80\%$ of the elitist solutions across generations. Nevertheless, CR-EA manage to adapt effectively, as reflected by the progression of marker colors, which transition from dark to bright and subsequently back to dark. This pattern indicates that the algorithm initially identifies the correct first-ranked solution, then experiences performance degradation due to noise effects (and potential Type~I errors), before eventually recovering and correctly identifying the first-ranked solution again.

Overall, the analysis highlights the effectiveness of the proposed novel method in enabling CR-EA to maintain a high-quality elite population under noisy conditions, demonstrating both robustness to noise and the capacity to recover from temporary ranking errors.

\subsection{Ablation and Parametric Study}
\label{sec: abla}

\begin{figure}[!ht]
  \includegraphics[width=0.48\textwidth]{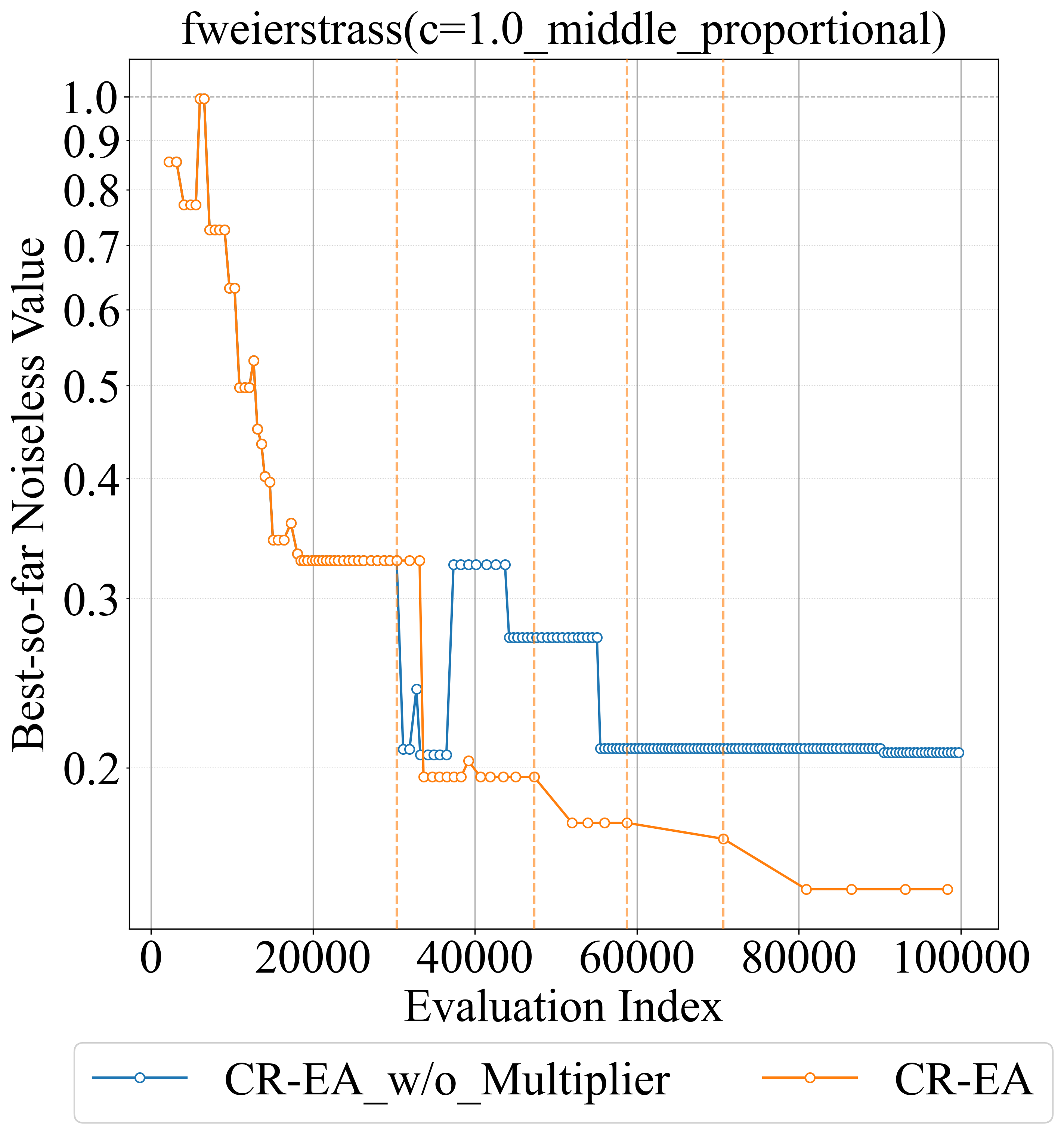}
  \caption{Ablation Study Convergence Plot}
  \label{fig:ablation}
\end{figure}

In Section \ref{sec: sampling budget update}, we discussed that the sampling budget is updated (multiplied by a factor $\alpha$) whenever the number of solutions tied with the $\gamma$-ranked solution exceeds $\beta N$. Here, we further justify the mechanism.  

To illustrate the effect of the sampling-budget update mechanism, an ablation study was conducted using two versions of the CR-EA algorithm (with $\gamma=1$) on fweierstrass problem (bounds between $[-100, 100]^{10}$) with identical random seeds. The only difference between the two is one includes the sampling-budget update mechanism, while the other does not. Figure~\ref{fig:ablation} presents the resulting convergence curves: the orange curve corresponds to the original algorithm, while the blue curve represents the version without the update mechanism. The vertical orange dashed line marks the point at which the update mechanism is triggered. As can be observed from Figure~\ref{fig:ablation}, CR-EA successfully triggers its sampling-budget update mechanism to avoid convergence stagnation, whereas the version without this mechanism reaches a convergence plateau.

The impact of the multiplier $\alpha$ is predictable as it directly relates to reducing the standard error of the sample mean by increasing the sample size. Although higher values of the multiplier $\alpha$ could also be used, they would increase computational expense which may not  provide proportionate benefits in accurate ranking. For this reason, and to avoid tuning to non-integer values of $\alpha$, we fixed $\alpha = 2$ in our experiments. This choice reduces the standard error of the sample mean by a factor of $\sqrt{2}$~\cite{sqrt_n_error}, and has consistently demonstrated competitive results in our experiments.
   
To determine the value of $\beta$ for CR-EA, a parameter study was conducted by running CR-EA with five candidate values: $0.05$, $0.1$, $0.2$, $0.5$, and $0.75$. The study focused on two representative problems: a unimodal problem (Benign Ellipsoid) and a multimodal problem (Ackley) where both are
bounded between $[-5, 5]^{10}$. 

Performance was evaluated by first identifying the global best and global worst objective values attained by any $\beta$ setting on each problem. Using these extrema, the \textit{score} of $\beta$ on a given problem is defined as:

\begin{equation}
\text{score} =
\frac{\text{global\_worst} - \text{\textit{average\_value}}}
{\text{global\_worst} - \text{global\_best}}
\label{eq:relative_score}
\end{equation}
Here, \textit{average\_value} denotes the average objective value across multiple runs 
for a given $\beta$ setting. This normalization places all algorithms on a common scale between 0 and~1, where higher values indicate better relative performance. 

Table~\ref{tab:beta_scores} reports the average scores across all test problems. 
The results indicate that CR-EA performs best for $\beta$ values between $0.2$ and $0.75$. However, larger values of $\beta$ delay activation of the sampling-budget update mechanism, resulting in significantly longer runtime before the evaluation budget is exhausted. Hence, $\beta = 0.2$ was selected as it offers a favorable balance between performance and computational time efficiency. 
The full results of the parameter study are provided in Section~II of \texttt{supplementary material.pdf}.

\begin{table}[!ht]
  \centering
  \caption{Average score for different values of $\beta$.}
  \label{tab:beta_scores}
  \begin{tabular}{c c}
    \toprule
    \textbf{$\beta$} & \textbf{Average Score} \\
    \midrule
    0.05 & 0.61 \\
    0.1  & 0.68 \\
    0.2  & 0.72 \\
    0.5  & 0.75 \\
    0.75 & 0.74 \\
    \bottomrule
  \end{tabular}
\end{table}

In addition to the experiments conducted on CR-EA, a parametric study was also performed for CR-CMA-ES-B. As discussed in Section~\ref{sec:method_cma}, the parameters $\gamma = 0.5N$ and $\beta = 0.5N$ were chosen to mirror the default CMA-ES parent selection mechanism, in which the top $0.5N$ ranked solutions are selected as parents. However, preliminary experiments showed that for the population size used in this study ($N=20$), this setting causes the sampling-budget adaptation mechanism to activate more frequently than in CR-EA. Consequently, using the same value of $\alpha = 2$ as in CR-EA may lead to overly aggressive increases in the global sampling budget $B_t$ for certain problem instances. To examine the influence of this parameter, a sensitivity analysis on $\alpha$ was conducted.

\begin{figure}[htb]
  \subfloat[Similar performance of $\alpha$ values.]
    {\includegraphics[width=0.48\textwidth]{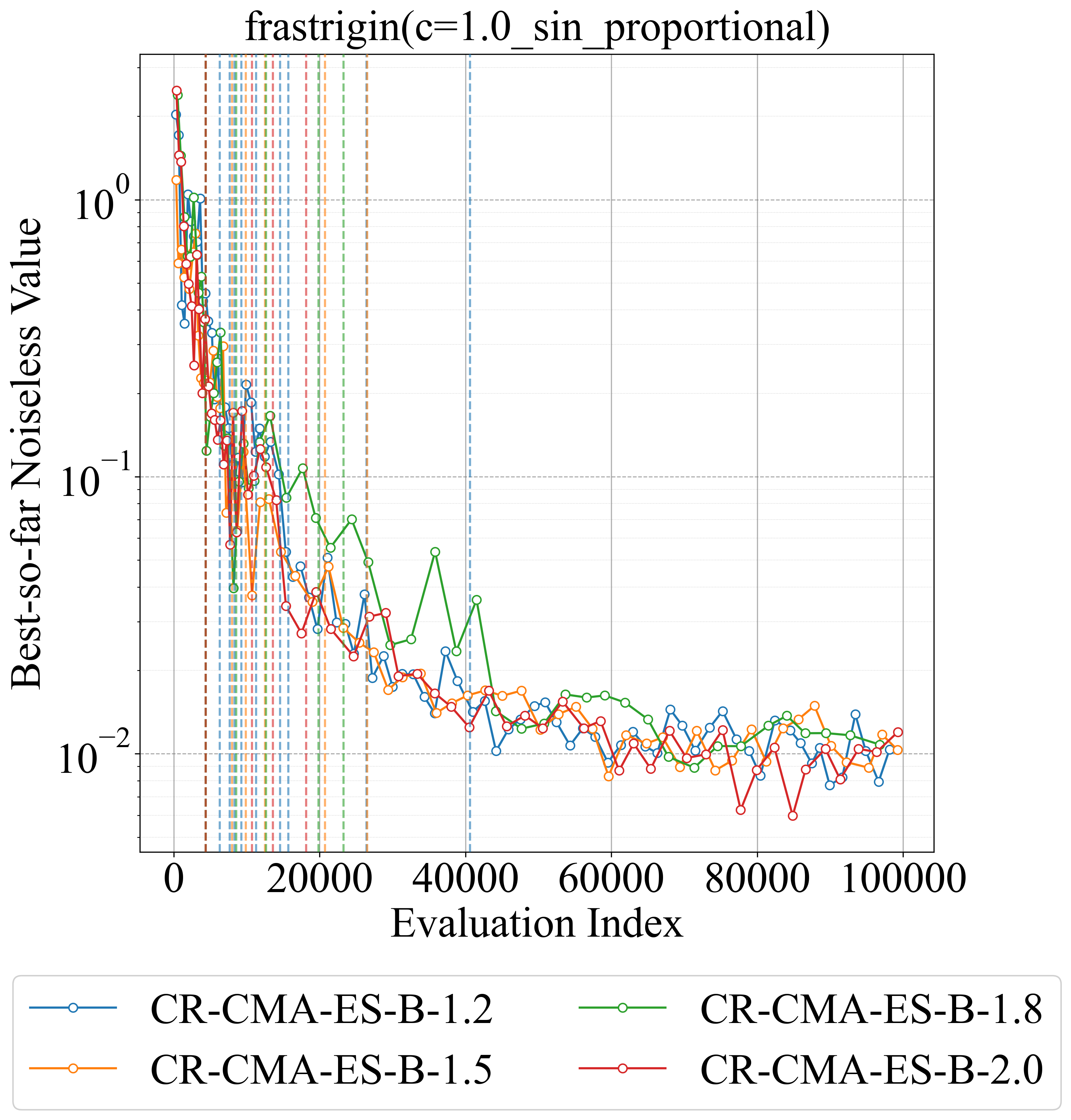}%
     \label{fig:abla_cma_1}}
      \hfil
  \subfloat[Better performance of lower $\alpha$ values.]
    {\includegraphics[width=0.48\textwidth]{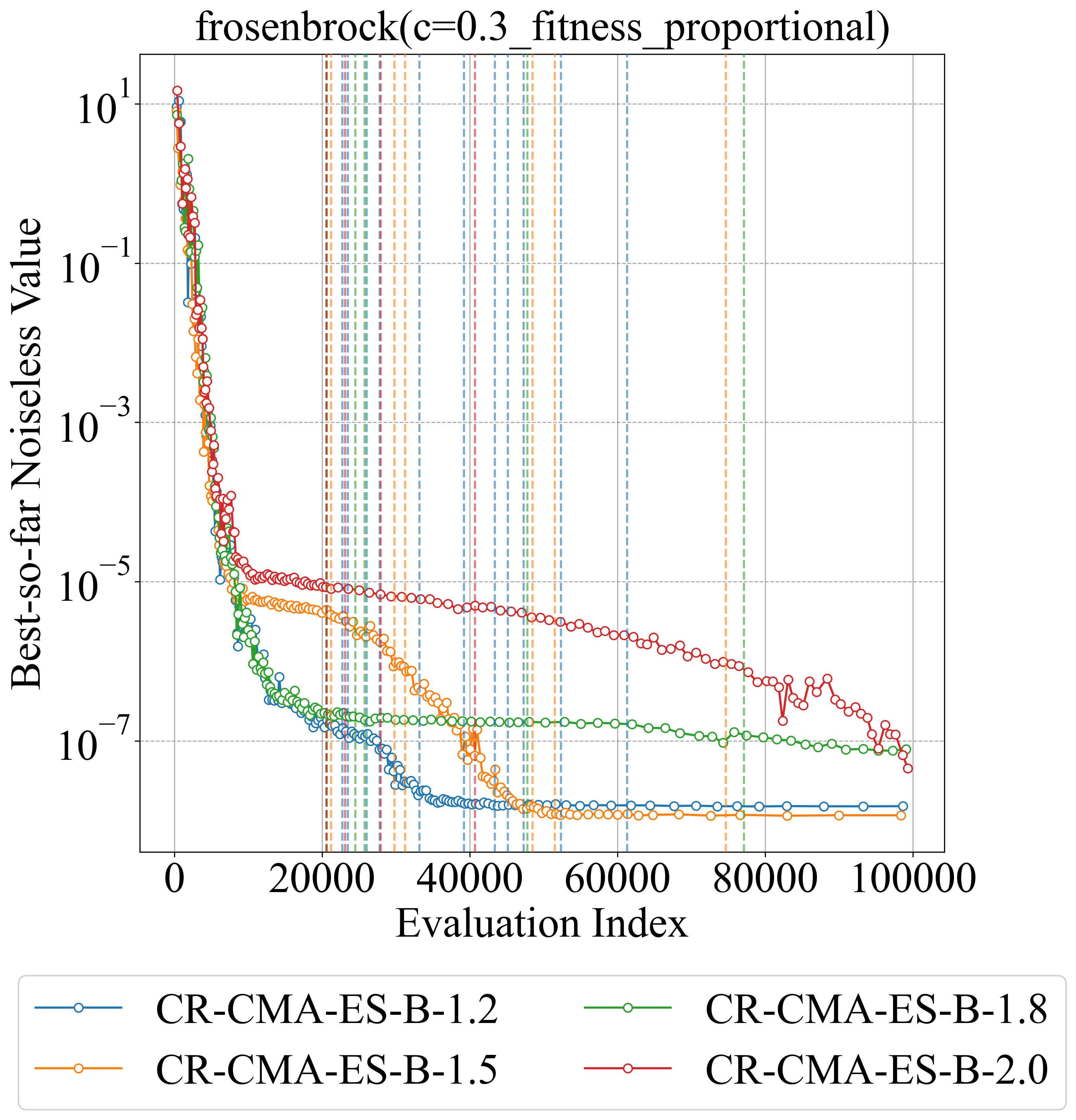}%
     \label{fig:abla_cma_2}}
     
  \subfloat[Worse performance of lower $\alpha$ values.]
    {\includegraphics[width=0.48\textwidth]{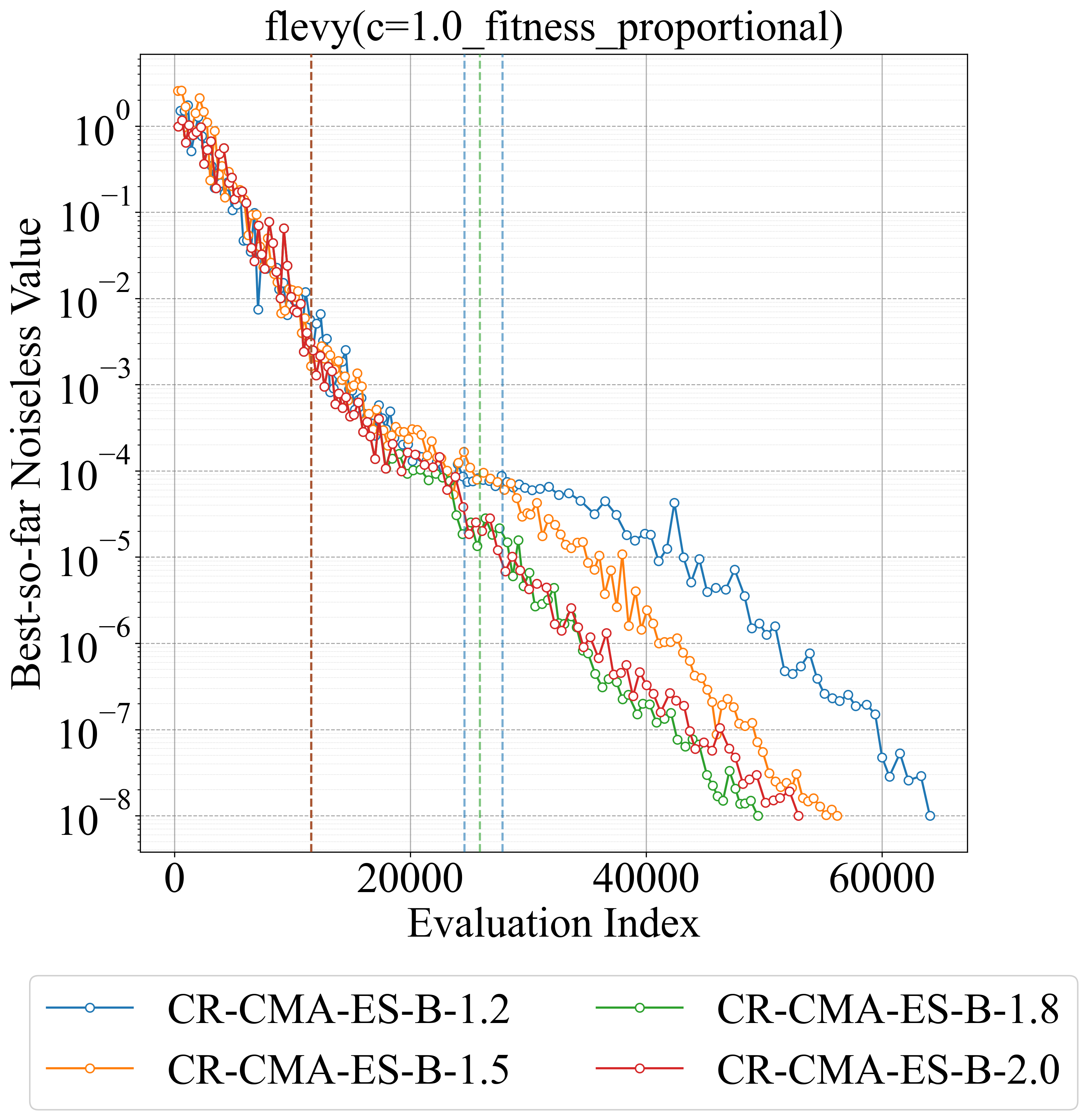}%
     \label{fig:abla_cma_3}}
      \hfil
  \subfloat[Better performance of $\alpha=2.0$.]
    {\includegraphics[width=0.48\textwidth]{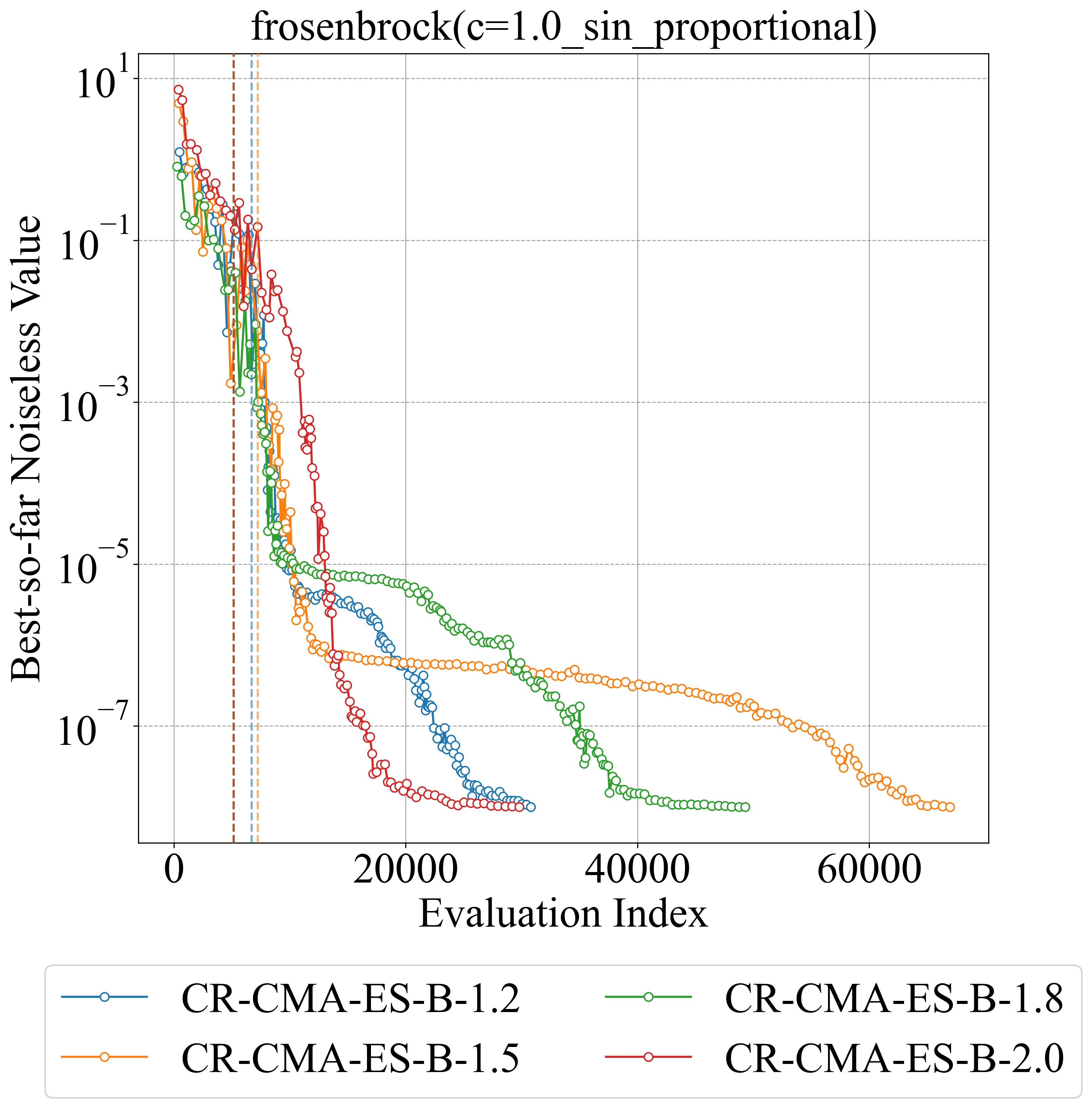}%
     \label{fig:abla_cma_4}}

  \caption{Parametric Study of CR-CMA-ES-B}
  \label{fig:abla_cma}
\end{figure}

CR-CMA-ES-B was evaluated using four values of $\alpha$: $1.2$, $1.5$, $1.8$, and $2.0$. The algorithm was tested across all problem--noise combinations described in Section~\ref{sec:benchmark problem} with 31 multiple runs. Representative results are shown in Figure~\ref{fig:abla_cma}, which illustrates the median run convergence behaviour for different values of $\alpha$, with the vertical dashed line indicating the point at which sampling-budget adaptation is triggered. The complete set of plots are provided in the \texttt{supplementary material.zip} file (folder: \textit{CR-CMA-ES Parametric Study}).

Overall, the results indicate that the algorithm is insensitive to the choice of $\alpha$, with all tested values producing similar convergence patterns in most scenarios (figure ~\ref{fig:abla_cma_1}). However, several cases exhibited improved performance when smaller values of $\alpha$ were used, as shown in figure~\ref{fig:abla_cma_2} and only one instance was observed in which a smaller value of $\alpha$ produced the worst performance (figure ~\ref{fig:abla_cma_3}). Nevertheless, there were also a few cases in which $\alpha = 2$ achieved the best result (figure ~\ref{fig:abla_cma_4}). Based on these findings, $\alpha = 1.2$ was selected for CR-CMA-ES-B, as it provided a more conservative parameter choice.

\section{Practical Applications}
\label{sec:olympus}
Having demonstrated performance on benchmark problems, we next evaluate the novel method on a set of simulated practical problems drawn from the Olympus framework~\cite{olympus}. Olympus is a chemistry-focused noisy-optimization platform that emulates experimental noise via a pretrained probabilistic Bayesian neural network (BNN), referred to as the experiment emulator. For our study, we selected two distinct emulators, summarized in Table~\ref{tab:olympus_table}. Each case corresponds to a continuous single-objective problem with clearly defined goals as documented in Olympus. 

We conducted the experiments using CMA-ES-B and EA algorithms on the Olympus benchmark suite and compare their performance. Following~\cite{olympus}, the evaluation budget is fixed at $10{,}000$ for all problems. The population size was set static to 20 for CMA-ES-based algorithms and to $10D$ for EA, where $D$ denotes the problem dimension. Each algorithm is executed 31 times per problem to ensure statistical reliability.

    \begin{table}[!ht]
      \centering
      \caption{Chosen Olympus experiment emulators}
      \label{tab:olympus_table}
      \begin{tabular}{l c c l}
        \toprule
        \textbf{Emulator}   & \textbf{Dim} & \textbf{Goal}   & \textbf{Description}                                      \\
        COLORS\_BOB        & 5                  & min        & Five dye mixture targeting normalized green-like RGB.     \\
        COLORS\_N9          & 3                  & min        & Three dye mixture targeting normalized green-like RGB.    \\
        \bottomrule
      \end{tabular}
    \end{table}

\begin{figure}[!htbp]
  \subfloat[CMA-ES-B on COLORS_BOB problem]
    {\includegraphics[width=0.48\textwidth]{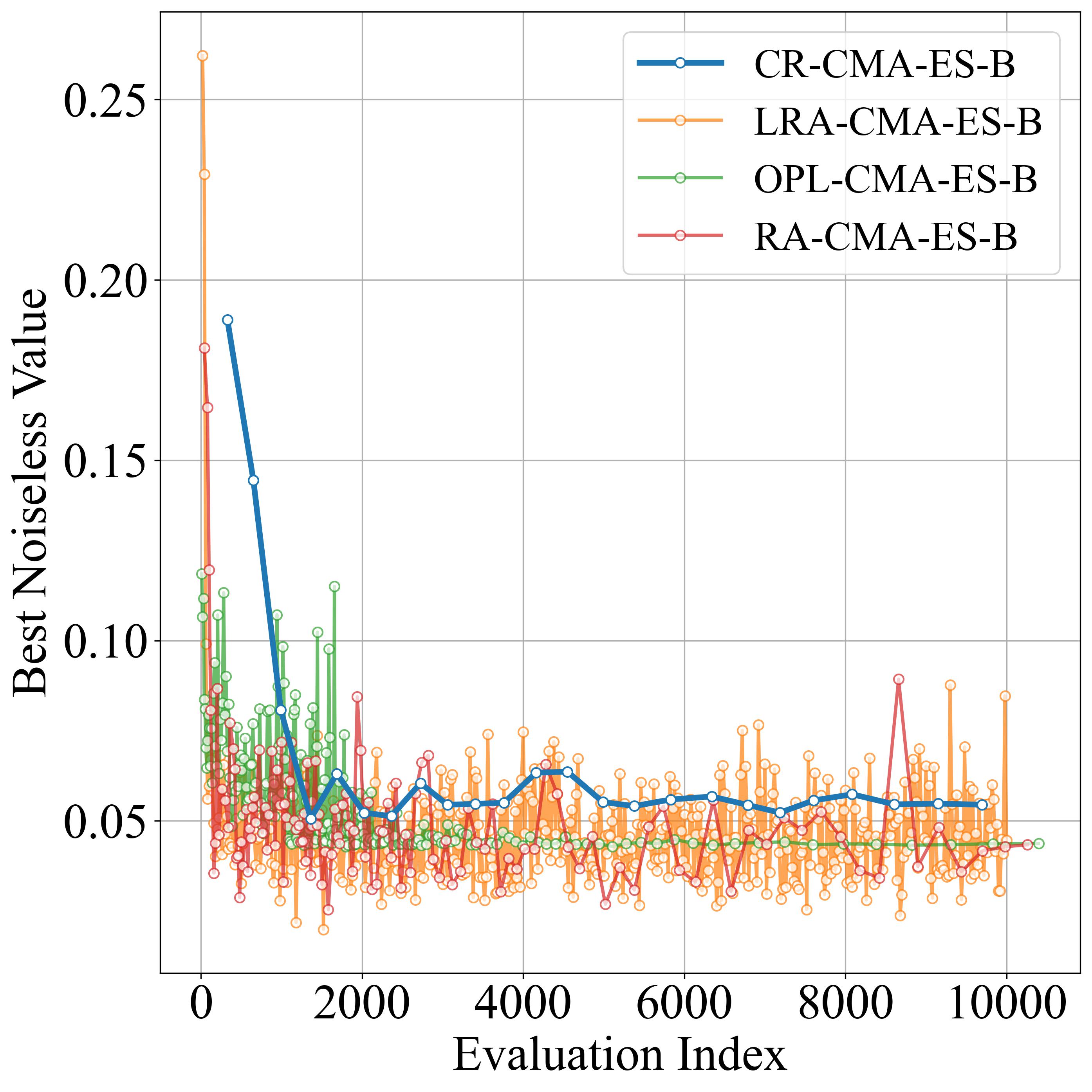}%
     \label{fig:olymp_plot_CMA-ES_1}}
  \hfil
  \subfloat[CMA-ES-B on COLORS_N9 problem]
    {\includegraphics[width=0.48\textwidth]{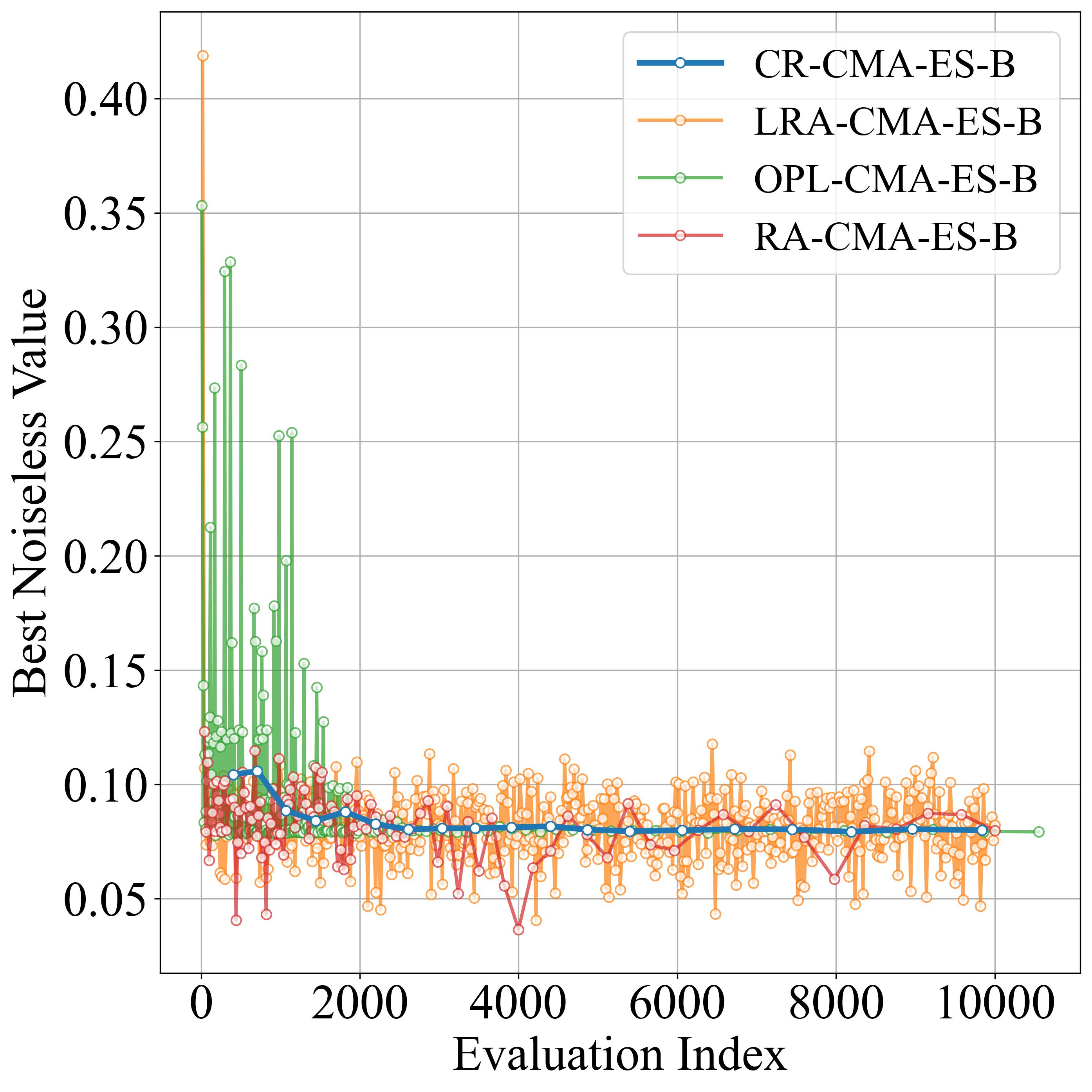}%
     \label{fig:olymp_plot_CMA-ES_2}}

  \caption{Convergence plots of CMA-ES variants on Olympus emulator.}
  \label{fig:olymp_plot_CMA-ES}
\end{figure}

\begin{figure}[!htbp]
  \subfloat[EA on COLORS_BOB problem]
    {\includegraphics[width=0.48\textwidth]{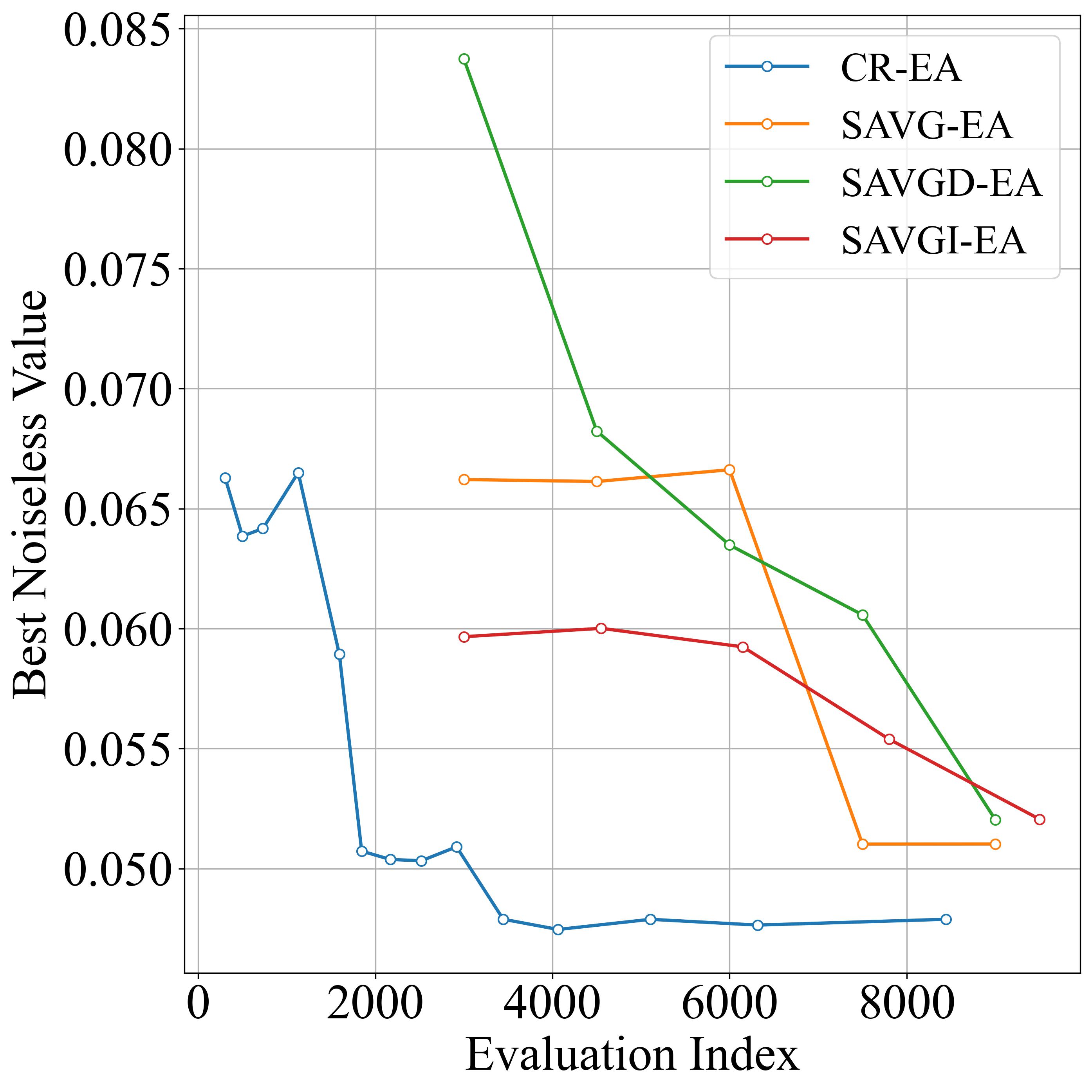}%
     \label{fig:olymp_plot_EA_1}}
  \hfil
  \subfloat[EA on COLORS_N9 problem]
    {\includegraphics[width=0.48\textwidth]{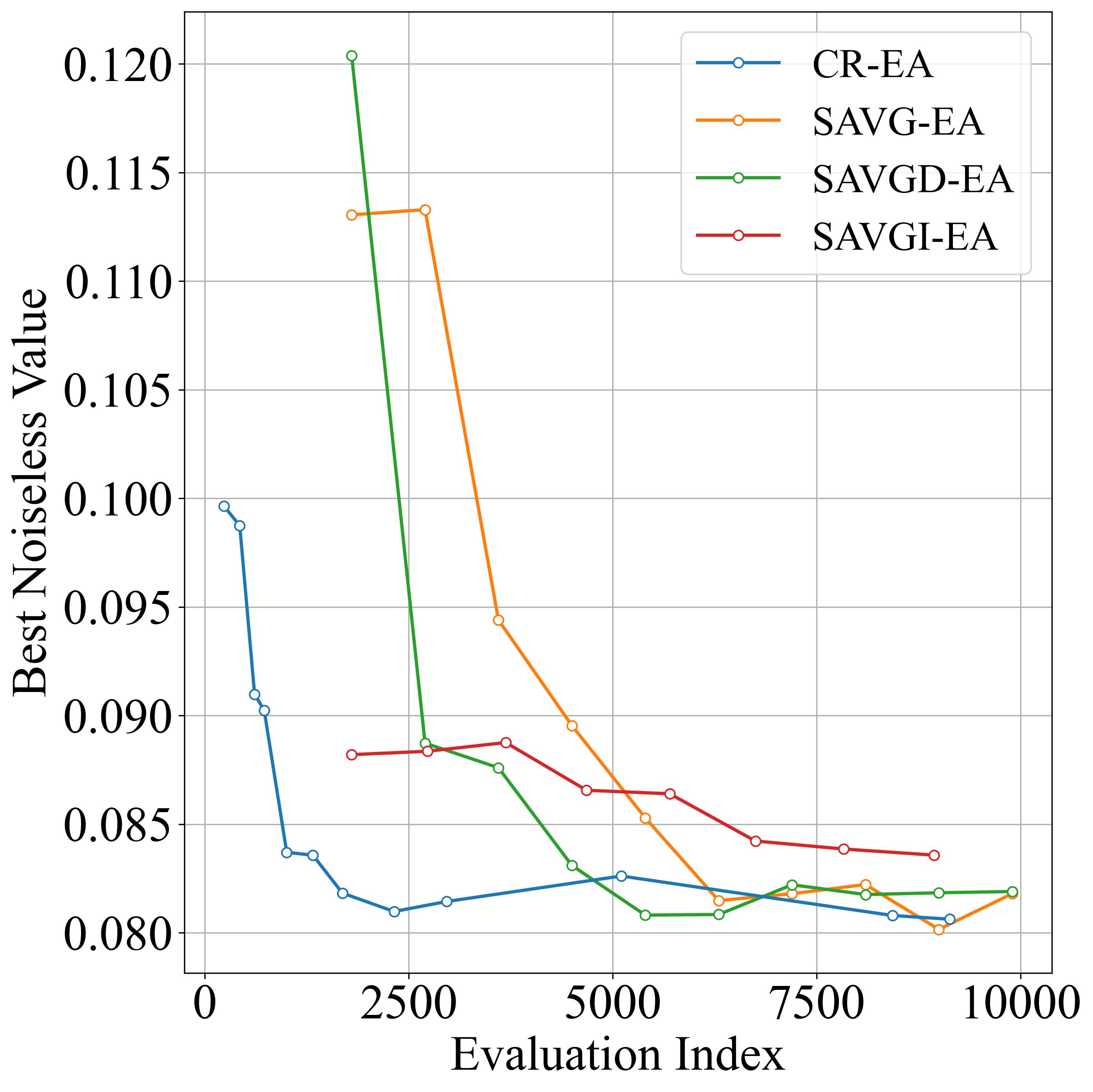}%
     \label{fig:olymp_plot_EA_2}}

  \caption{Convergence plots of EA variants on Olympus emulator.}
  \label{fig:olymp_plot_EA}
\end{figure}

Figures~\ref{fig:olymp_plot_CMA-ES} and \ref{fig:olymp_plot_EA} presents the median-run convergence plots of CMA-ES and EA respectively. To estimate the true mean objective value, each proposed solution was resampled 1,000 times in post-processing. 

As evident from Figure ~\ref{fig:olymp_plot_CMA-ES}, CR-CMA-ES performance is competitive on both problems and posses significantly lower volatility compared to other CMA-ES approaches. In addition, Figure \ref{fig:olymp_plot_EA} shows that CR-EA converges faster on both problem and manage to obtain superior result on COLORS_BOB problem (figure \ref{fig:olymp_plot_EA_1}. These results further reinforces the proposed novel method's effectiveness, particularly in black-box simulation scenarios with unknown noise profiles. 
        
\section{Conclusion}
\label{sec:conc}
 
In this paper, we introduced a novel confidence ranking (CR) method with adaptive sampling budget allocation for noisy black-box optimization. The method incorporates an explicit resampling strategy that allocates additional evaluations to the most promising solutions, together with a confidence-based ranking mechanism using Welch’s $t$-test. The method is then integrated into CMA-ES and EA framework. As a result, the algorithms are able to effectively handle both homoscedastic and heteroscedastic noise without requiring prior knowledge of problem-specific features. 

Extensive experiments were conducted on five unimodal and five multimodal benchmark problems, each evaluated under five distinct noise models and two noise levels. The results demonstrate that the algorithms utilizing the novel method consistently show superior performance compared to others. Furthermore, evaluations on a chemistry-focused emulators (Olympus) confirm that the method is reliable under realistic experimental noise, further validating its practical utility. 

Beyond these promising results, several research directions could further enhance this work. First, the method's reliance on explicit averaging provides a rigorous statistical foundation for embedding surrogate models directly, offering a clear path to reduce expensive function evaluations. Second, we plan to extend CR-EA to constrained noisy optimization and noisy multi-objective optimization, broadening its applicability to a wider range of real-world engineering and scientific problems under uncertainty. Finally, extensions to more than one objective could also be considered.

The source code is available from the authors upon request and will be made publicly accessible following the publication of the paper.

\section*{Acknowledgment}
Generative AI tools were used to assist with improving the clarity, readability, and expression of the writing of this work. 

\bibliographystyle{ACM-Reference-Format}
\bibliography{refs}   

\end{document}